\PassOptionsToPackage{table}{xcolor}%
\documentclass{article} %
\usepackage{iclr2027_conference,times}

\iclrfinalcopy
\AtBeginDocument{\let\origmaketitle\maketitle\renewcommand{\maketitle}{\origmaketitle\lhead{Preprint}}}
\newcommand{\aff}[1]{\textsuperscript{\mdseries #1}}
\newcommand{\affb}[1]{\textsuperscript{\bfseries #1}}%
\newcommand{\authsep}{\hspace{1.5em}}

\usepackage{amsmath,amsfonts,bm}

\def\eqref#1{equation~\ref{#1}}

\def\1{\bm{1}}

\DeclareMathAlphabet{\mathsfit}{\encodingdefault}{\sfdefault}{m}{sl}
\SetMathAlphabet{\mathsfit}{bold}{\encodingdefault}{\sfdefault}{bx}{n}

\usepackage{hyperref}
\usepackage{url}
\usepackage{fontawesome5}
\newcommand{\hficon}{\raisebox{-0.55ex}{\includegraphics[height=1.35em]{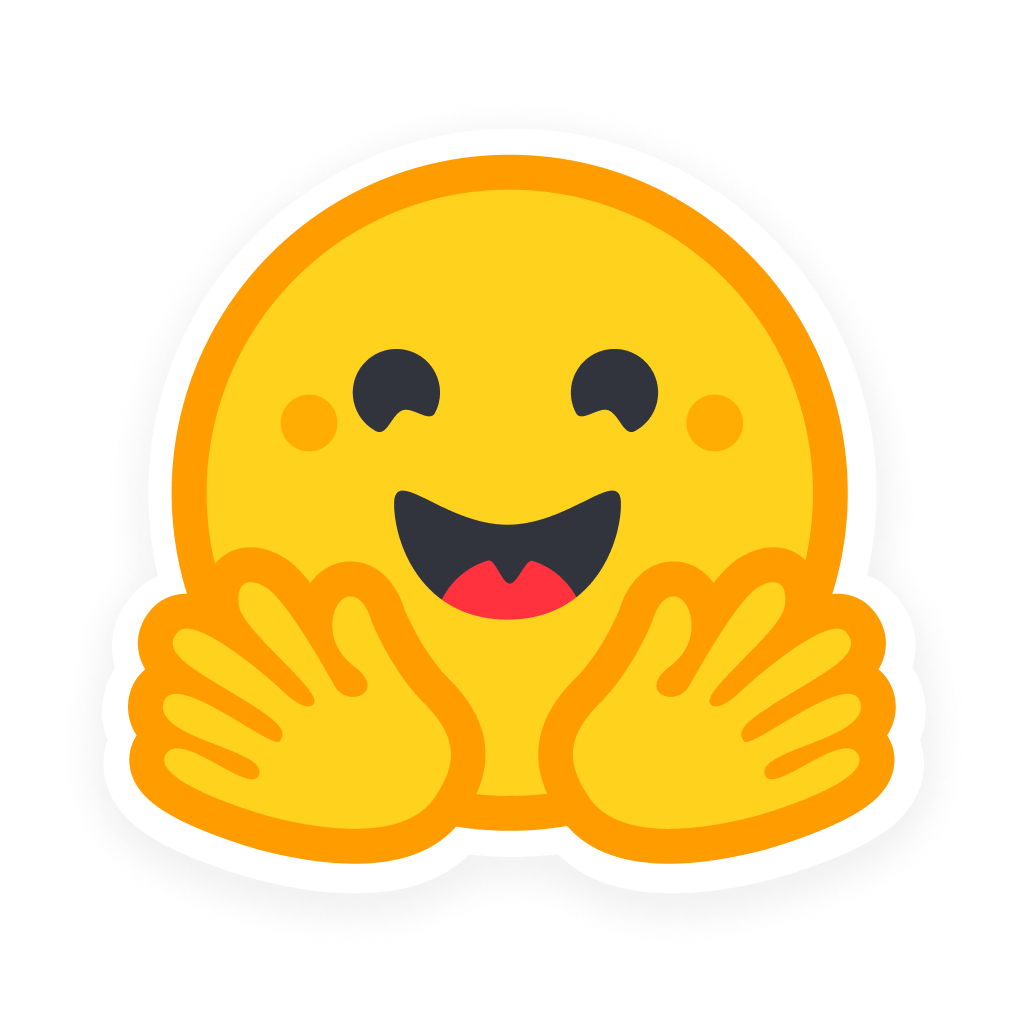}}}
\newcommand{\ghicon}{\raisebox{-0.3ex}{\scalebox{1.25}{\faGithub}}}

\usepackage{booktabs}

\usepackage[most]{tcolorbox}
\usepackage{enumitem}
\usepackage{xcolor}
\usepackage{graphicx}
\usepackage{pgfplots}
\pgfplotsset{compat=1.18}

\usepackage[framemethod=TikZ]{mdframed}

\usepackage{amssymb}
\usepackage{bbm}
\usepackage{booktabs}
\usepackage{multirow}

\usepackage{wrapfig}   %
\usepackage{needspace}
\usepackage{makecell}

\usepackage{soul}
\usepackage[capitalise,noabbrev,nameinlink]{cleveref}

\newcommand{\up}{\,{\scriptsize\textcolor{blue!65!black}{$\uparrow$}}}
\newcommand{\dn}{\,{\scriptsize\textcolor{red!75!black}{$\downarrow$}}}
\newcommand{\nc}{\,{\scriptsize\textcolor{gray}{$=$}}}

\title{Learning \emph{What} vs.\ Learning \emph{How}: \\ Decomposing Task and Interface Competence in Agentic Training}

\author{%
\textbf{Chaeeun Lee}\affb{1,4,*,\textdagger} \authsep
\textbf{Hossein A.~Rahmani}\affb{2,4,*,\textdagger} \authsep
\textbf{Aryo Pradipta Gema}\affb{1} \authsep
\textbf{Kexin Xu}\affb{3,4,\textdagger} \\
\textbf{Emine Yilmaz}\affb{2} \authsep
\textbf{Pasquale Minervini}\affb{1,5} \authsep
\textbf{T.~Ian Simpson}\affb{1} \\
\textbf{Chen Jin}\affb{4} \authsep
\textbf{Amrutha Saseendran}\affb{4} \\[0.6em]
\aff{1}School of Informatics, University of Edinburgh \quad
\aff{2}Centre for AI, University College London \\
\aff{3}Weatherall Institute of Molecular Medicine, University of Oxford \\
\aff{4}AI for Science \& Innovation, Enterprise AI, AstraZeneca \quad 
\aff{5}Miniml.AI \\[0.3em]
}

\begin{document}

\maketitle
{\renewcommand{\thefootnote}{\fnsymbol{footnote}}\footnotetext[1]{Equal contribution.}\footnotetext[2]{Work done during internship at AstraZeneca.}}

\begin{abstract}
LLM agents operate through \emph{harnesses} that define system prompts, action formats, feedback loops, and resource constraints.
A model may possess the knowledge needed to solve a task yet fail to express that knowledge through a particular harness.
This raises a basic question: when a policy fails under a given harness, is it missing task knowledge or the competence to operate the interface?
We distinguish two components of agentic competence: \emph{task competence}, the knowledge and reasoning needed to solve a task, and \emph{interface competence}, the ability to act on that knowledge through a given harness.
Across four reasoning and agentic benchmarks, we show that harness-conditioned training yields vastly different outcomes depending on how these interact.
Multi-harness distillation aggregates task knowledge from trajectories collected through different harnesses into a shared policy, but this knowledge is realised only when training preserves or improves competence on the inference-time interface.
Otherwise, single-harness training can induce \emph{interference} in interface competence, sometimes catastrophically collapsing performance on a held-out harness despite improvements in underlying task knowledge.
To disentangle task and interface supervision, we introduce \emph{harness rehearsal}: distillation on traces that carry minimal task knowledge while exercising the target interface.
Cross-harness task supervision or rehearsal alone yields little gain, but their combination yields gains far larger than either: on BFCL V3 multi-turn with Qwen3-4B, they achieve 8.5\% and 32.5\% respectively, versus 58.8\% together.
Finally, on-policy reinforcement learning (RL) is more robust than distillation to single-harness training: it does not substantially degrade performance on other harnesses, and interface competence acquired through multi-harness distillation persists through subsequent RL.
\end{abstract}
\begin{center}\vspace{-8pt}\small
\href{https://github.com/chaeeunlee-io/harness-competence}{\ghicon\ \textbf{\texttt{GitHub}}}\hspace{3em}
\href{https://huggingface.co/datasets/chaeeunlee/harness-competence}{\hficon\ \textbf{\texttt{HuggingFace}}}
\end{center}

\section{Introduction}
\label{sec:intro}
\setlength{\textfloatsep}{13pt}
\begin{figure}[t]
\centering
\includegraphics[width=\textwidth]{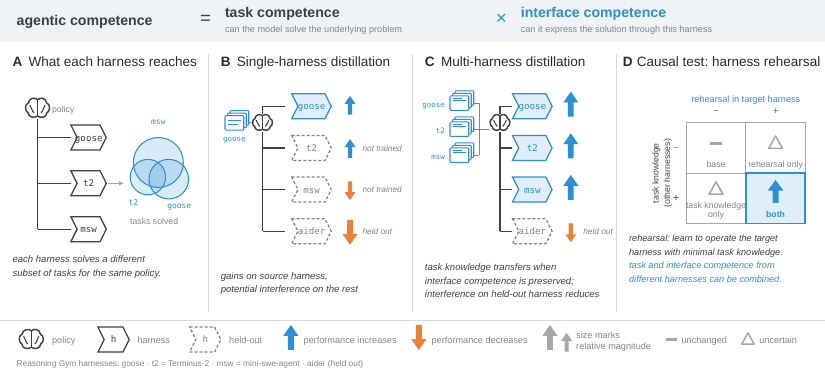}
\caption{\textbf{Decomposing task and interface competence.}
\textbf{(A)} A fixed policy shows different capabilities across harnesses.
\textbf{(B)} Single-harness distillation improves performance on the training harness but can degrade performance on others.
\textbf{(C)} Multi-harness distillation further improves on training harnesses and reduces performance degradation on held-out harnesses.
\textbf{(D)} Harness rehearsal provides interface supervision without requiring the model to solve the underlying task, while cross-harness distillation provides successful task demonstrations without exposing the model to the target harness.}
\label{fig:hero}
\end{figure}

Language model agents are defined not only by what they know, but by what they can do: use tools, execute code, and complete tasks over long interaction horizons~\citep{DBLP:conf/iclr/YaoZYDSN023,DBLP:conf/nips/SchickDDRLHZCS23,DBLP:conf/icml/WangCY0L0J24}.
In these settings, an agent acts through a \emph{harness}, an interface that determines what it can observe, which actions it can take, and how those actions are interpreted~\citep{DBLP:journals/corr/abs-2605-27922,DBLP:journals/corr/abs-2607-28802,DBLP:journals/corr/abs-2605-22166}.
This creates a consequential source of failure: a policy may know how to solve a task yet fail because it cannot express that solution through a given harness.
A coding agent may identify the correct fix but mishandle the shell.
A clinical reasoning agent may reason correctly about a patient record but fail to retrieve or submit the required information through the available interface.
Observed performance under a harness, therefore, reflects both the quality of reasoning about the task and the ability to carry out the intended actions through the harness.
We ask whether this distinction is merely descriptive or whether agentic competence can be meaningfully decomposed into components that can be learned, transferred, and lost independently.
We distinguish two components of agentic competence (\cref{fig:hero}): \emph{task competence}, the knowledge and reasoning needed to solve a task, and \emph{interface competence}, the ability to carry out a solution through a particular harness.
We hypothesise that task competence can transfer across harnesses, but expressing it through a given harness depends on interface competence with that harness.
We test this hypothesis through four questions. \textbf{Q1. Does harness choice only affect overall performance, or do different harnesses expose different aspects of a policy’s capabilities?} We find that harnesses differ not only in overall performance but also in which tasks they enable the policy to solve: given a frozen model, weaker harnesses can expose capabilities that stronger harnesses do not (\cref{sec:res_exposure}).
\textbf{Q2. How does supervision from one or more harnesses transfer to other harnesses?} Distilling on successful trajectories from one harness improves performance on the training harness but can degrade performance on others, in some cases to near zero.
Multi-harness distillation improves performance across the training harnesses while largely avoiding this collapse on a held-out harness (\cref{sec:res_ladder}). \textbf{Q3. Do task and interface supervision have independent effects, and how do they interact?} We test the task--interface competence decomposition and probe the contribution of each component by combining distillation traces from other harnesses with harness rehearsal on the target interface. Together, they improve target-harness performance far more than either alone, directly testing the interaction between the two sources of supervision (\cref{sec:res_drills}).
\textbf{Q4. After multi-harness distillation, how do the effects of single- and multi-harness RL compare with those of distillation?} Subsequent RL exhibits a different cross-harness pattern from distillation: single-harness RL does not produce the cross-harness degradation seen with single-harness supervised fine-tuning (SFT), and can sometimes improve performance on other harnesses, while held-out harness performance remains stable. Results from both the base model and the multi-harness distillation checkpoint suggest that GRPO is generally more robust to single-harness training than distillation. Consequently, multi-harness interface competence acquired during distillation can be carried through the subsequent RL stage, motivating how training harnesses can be best incorporated into the conventional SFT-then-RL post-training regime (\cref{sec:res_rl}).
\section{Related Work}

Harnesses are becoming an increasingly important part of agentic systems~\citep{DBLP:conf/nips/YangJWLYNP24,DBLP:conf/iclr/0001LSXTZPSLSTL25,DBLP:conf/icml/WangCY0L0J24}, shaping not only how agents are evaluated~\citep{DBLP:journals/corr/abs-2603-13966,DBLP:journals/corr/abs-2605-27922,DBLP:conf/iclr/MerrillSCLRBSSW26} but also how they are trained~\citep{DBLP:journals/corr/abs-2609-04518,DBLP:journals/corr/abs-2605-08741}. Recent work has begun to study this from several angles. Harness-Bench systematically varies harness configurations to quantify their effect on agent performance \citep{DBLP:journals/corr/abs-2605-27922}. Other work asks how the harness itself can be improved: Meta-Harness and HarnessOpt-Bench study harness optimisation, while HarnessDev and Self-Harness explore agents that create or refine their own harnesses \citep{DBLP:journals/corr/abs-2603-28052,DBLP:journals/corr/abs-2608-06301,DBLP:journals/corr/abs-2609-01437,DBLP:journals/corr/abs-2606-09498}.

Rather than adapting the harness to the policy, another line of work studies how policies learn when training incorporates multiple harnesses. HarnessBandit studies scheduling across harnesses for agentic RL \citep{DBLP:journals/corr/abs-2609-13739}, \emph{What Does Multi-Harness RL Learn?} examines cross-harness credit assignment and portability \citep{DBLP:journals/corr/abs-2609-04518}, as does HarnessX \citep{DBLP:journals/corr/abs-2606-14249}, and \emph{Co-Evolving Harnesses and Models} jointly evolves harnesses and policies \citep{DBLP:journals/corr/abs-2609-09134}. These works establish transfer across harnesses as an explicit training problem~\citep{DBLP:journals/corr/abs-2609-13739,DBLP:journals/corr/abs-2609-04518}.

Our work focuses on what determines this transfer. We separate task supervision from learning to operate the target inference-time interface, allowing us to study when supervision collected through one harness improves performance through another, when interface-specific learning helps or interferes with transfer, and whether the benefits of multi-harness distillation persist through the subsequent RL stage.

\begin{figure}[t]
    \centering
    \includegraphics[width=\textwidth]{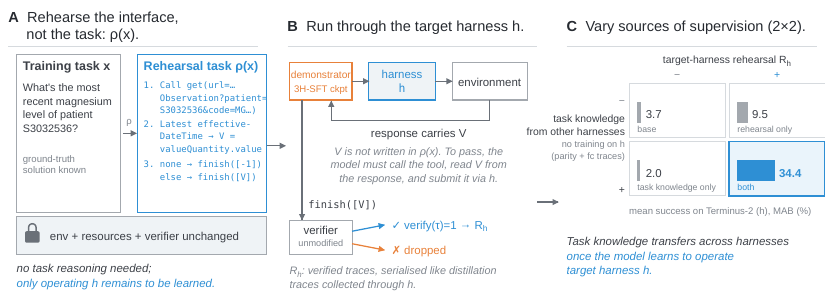}
    \caption{\textbf{Harness rehearsal.} (A) Replace the task instruction with $\rho(x)$, which specifies the correct actions while preserving the environment and verifier. (B) Successful traces through harness $h$ form the rehearsal set $\mathcal{R}_h$. (C) Crossing rehearsal with task supervision from other harnesses gives a $2{\times}2$ that separates task supervision from interface supervision.}
    \label{fig:harness_rehearsal}
\end{figure}

\section{Decomposing Task and Interface Competence} \label{sec:framework}

Our goal is to understand how training through agent harnesses shapes a shared policy's task and interface competence, and how the two contribute to performance across harnesses. We hypothesise that task competence can transfer across harnesses, but realising it depends on interface competence with the harness used at inference time. We test this hypothesis through four experiments, corresponding to the four questions introduced in \cref{sec:intro}. For Q1, we compare a frozen policy across harnesses (\cref{sec:method_framework}). In \cref{sec:method-training}, we distill the student on successful trajectories from one or multiple training harnesses and measure cross-harness transfer and interference (Q2). In \cref{sec:method_rehearsal}, we decouple task and interface supervision: cross-harness distillation provides task supervision, while harness rehearsal provides interface supervision on the target harness, allowing us to test how the two interact (Q3). Finally, in \cref{sec:method_rl}, we continue multi-harness SFT checkpoints with Group Relative Policy Optimization (GRPO) using one or several harnesses and compare its cross-harness effects with those from distillation (Q4).

\subsection{The Harness as Training and Evaluation Environment}
\label{sec:method_framework}

\begin{wrapfigure}[26]{r}{0.52\textwidth}
\vspace{-35pt}
\centering
\includegraphics[width=0.52\textwidth]{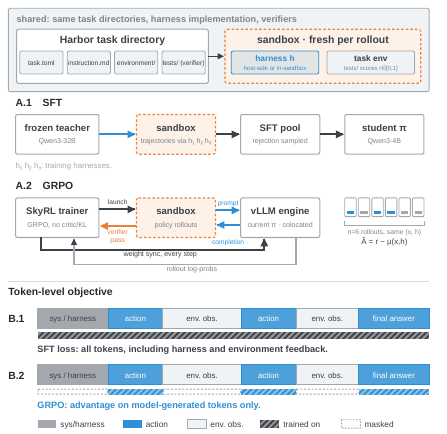}
\caption{\textbf{Shared sandboxed task execution for SFT and RL.} Harbor runs each task in an isolated container with the same harness code and verifier for SFT trace collection (A.1), GRPO rollouts (A.2), and evaluation. The SFT loss applies to all trajectory tokens (B.1), while GRPO applies advantage only to policy-generated tokens (B.2).}
\label{fig:framework}
\end{wrapfigure}

All experiments in this study use Harbor~\citep{team2026harbor} as the common execution framework for SFT data collection, RL rollouts, and evaluation (\cref{fig:framework}). Each task specifies an instruction, environment image, configurations, and verifier, and is executed in an isolated container. For GRPO, SkyRL~\citep{DBLP:journals/corr/abs-2511-16108} operates on Harbor's rollout loop, querying a colocated inference server and broadcasting updated weights between steps to keep rollouts on-policy.

For Q1, we evaluate the frozen base policy through every harness on the same tasks. We compare overall success rates and overlap in the tasks solved across harnesses, asking both how strongly performance depends on the interface and whether different harnesses expose different capabilities of the same policy. For each benchmark, the harness set $\mathcal H$ contains three training harnesses $\mathcal H_{\mathrm{train}}$ and one held-out harness $h_{\mathrm{held}}$, which is excluded from training. A policy $\pi$ acting on task $x$ through harness $h$ produces a trajectory $\tau\sim\pi(\cdot\mid x,h)$, scored by the task verifier as $s(\tau)\in[0,1]$. We define success as $r(\tau)=\mathbf{1}[s(\tau)\geq 0.99]$. The verifier score is already binary on MedAgentBench, AppWorld, and BFCL. Reasoning Gym grades a submission against the generator that produced the task, so the threshold matters only there.

\subsection{Multi-Harness Distillation}
\label{sec:method-training}

To answer Q2, we compare how single- and multi-harness distillation affect performance across harnesses. For each $h \in \mathcal{H}_{\mathrm{train}}$ we sample teacher trajectories on the training tasks and keep only those with $r(\tau)=1$, giving a distillation dataset $\mathcal{T}_h$ (training settings in Appendix \cref{tab:sft_hparams}, trace collection settings in Appendix \cref{tab:pool_inventory}). A student is trained either on single-harness distillation traces or on the multi-harness union of traces, $\mathcal{T}_{\mathrm{MH}} = \bigcup_{h\in\mathcal{H}_{\mathrm{train}}}\mathcal{T}_h$,
with the objective
\begin{equation}
\mathcal{L}(\pi)
= -\mathbb{E}_{\tau\sim\mathcal{T}}
\textstyle\sum_{t}
\log \pi(x_t \mid x_{<t}),
\label{eq:mhd_objective}
\end{equation}
where $\mathcal{T}$ is the chosen distillation set, and the objective is computed over every token $x_t$ in the serialised trajectory. The loss covers the full trajectory, including harness-specific interaction and feedback.

We refer to a policy distilled on a single harness as a \emph{specialist}, and abbreviate the policy distilled on three training harnesses as \emph{3H}. Relative to the initial policy, we call an increase in performance on a harness (positive) \emph{transfer} and a decrease \emph{interference}. We evaluate every trained policy on all training harnesses and on at least one held-out harness. Comparing specialist policies with the 3H policy lets us assess whether multi-harness distillation mitigates the cross-harness interference seen with single-harness training. We also examine whether multi-harness distillation yields larger gains than single-harness distillation, suggesting transfer of task knowledge learned through other harnesses. We test our hypothesis more directly with harness rehearsal in \cref{sec:method_rehearsal}.

\subsection{Harness Rehearsal}
\label{sec:method_rehearsal}

Q3 asks whether task and interface supervision have independent effects, and how they interact. We introduce \textbf{harness rehearsal} (\cref{fig:harness_rehearsal}), which provides target-interface supervision while minimising supervision on how to reason about and solve the underlying task. The solution is given in the instruction, leaving the model only to execute it through the target harness. Here, the \emph{target harness} is the harness the model is evaluated on. In harness rehearsal experiments, rehearsal data are generated for the target harness. Solving an original training instance requires both determining which operations solve the task and executing those operations through a given harness. Harness rehearsal removes the former requirement: for each \(x \in \mathcal{D}_{\mathrm{train}}\) with ground-truth solution \(y\), we programmatically construct an instruction \(\rho(x)\) that specifies the operations required to obtain and submit \(y\), including any values that must be extracted from intermediate results. The environment, resources, and verifier are unchanged, so the model has to execute the prescribed operations correctly through the target harness \(h\) to pass the verifier. We collect rehearsal trajectories by running 3H checkpoint as the demonstrator policy on these instructions through \(h\) and retaining trajectories that pass the original verifier:
\begin{equation}
\mathcal{R}_h
=
\left\{
\tau :
\tau \sim \pi_{\mathrm{demo}} \text{ on } (\rho(x), h),\;
x \in \mathcal{D}_{\mathrm{train}},\;
r(\tau)=1
\right\}.
\end{equation}
We train on the resulting $\mathcal{R}_h$ with the full-trajectory SFT objective in \cref{eq:mhd_objective}. An abridged example of a BFCL V3 multi-turn rehearsal instruction for the \textsc{fc} target harness reads:

\begin{mdframed}[
    linecolor=gray!70,
    linewidth=2.5pt,
    topline=false,
    rightline=false,
    bottomline=false,
    backgroundcolor=gray!8,
    innertopmargin=8pt,
    innerbottommargin=8pt,
    innerleftmargin=12pt,
    innerrightmargin=12pt
]
\small
\begin{enumerate}[leftmargin=*, itemsep=4pt]
    \item Call \texttt{ls(a=True)} and read its result.

    \item Let $V$ be the \textbf{first} entry listed by that result
    which does not start with a dot. Call
    \texttt{touch(file\_name='rehearsal\_<V>.txt')}, substituting $V$.
\end{enumerate}
\end{mdframed}

We combine harness rehearsal with \emph{cross-harness distillation} in a \(2\times2\) design. This experimental setup allows us to measure their individual effects and test whether target-harness experience improves task knowledge transfer from cross-harness distillation (\cref{tab:res_drills}). We call supervision \emph{cross-harness} when it is generated using a harness other than the target harness. Cross-harness distillation provides successful task demonstrations without exposure to the evaluation interface, while harness rehearsal provides target-interface experience without requiring the model to reason about and determine the solution.

\subsection{Multi-Harness Reinforcement Learning}
\label{sec:method_rl}

Q4 examines how RL reshapes competence acquired through multi-harness SFT, including whether single-harness RL reproduces the cross-harness interference observed under specialist distillation. We apply GRPO~\citep{DBLP:journals/corr/abs-2402-03300} to SFT checkpoints using the same training harnesses, task environments, and verifiers as in distillation. For each task \(x\) and harness \(h \in \mathcal{H}_{\mathrm{train}}\), we sample \(G\) rollouts \(\tau_{x,h,1},\ldots,\tau_{x,h,G} \sim \pi_\theta(\cdot \mid x,h)\) with binary verifier reward \(r(\tau)\in\{0,1\}\), and compute
\begin{equation}
\hat{A}_{x,h,i}
=
r(\tau_{x,h,i})
-
\frac{1}{G}\sum_{j=1}^{G} r(\tau_{x,h,j}).
\label{eq:grpo_adv}
\end{equation}
Advantages are computed within each task--harness group and applied only to policy-generated tokens. We omit the standard-deviation normalisation of the original estimator, which inflates advantages for low-variance groups~\citep{DBLP:journals/corr/abs-2503-20783}. Mean-centering also renders groups with identical rewards exactly neutral. We use \(G=6\), no additional reward shaping, and no KL penalty. Full hyperparameters are given in Appendix \cref{tab:grpo_hparams}. We apply GRPO on a fixed \emph{learning band}: training tasks for which the initial checkpoint succeeds on some attempts but not all on at least one training harness (\cref{app:rl}). This also concentrates rollouts on tasks where reward variation is more likely. The experiment therefore focuses on how RL changes existing task and interface competence, rather than whether it can solve previously unsolved tasks. We compare multi-harness and single-harness GRPO, matching total rollout compute by increasing the number of epochs in the single-harness setup (Appendix \cref{tab:rl_budgets}).

\section{Experimental Settings}
\label{sec:exp_settings}

\paragraph{Benchmarks.}
We use four benchmarks spanning different forms of reasoning and agent--environment interaction: MedAgentBench (MAB) \citep{doi:10.1056/AIdbp2500144}, Reasoning Gym (RG) \citep{DBLP:conf/nips/StojanovskiSSJA25}, AppWorld \citep{DBLP:conf/acl/TrivediKHMDLGSB24}, and the Berkeley Function-Calling Leaderboard (BFCL V3) \citep{DBLP:conf/icml/PatilMYJSSG25}. MAB and AppWorld involve stateful environments, RG consists of programmatically generated reasoning tasks, and BFCL evaluates single- and multi-turn function calling (BFCL-ST and BFCL-MT). Training and test tasks are disjoint throughout. MAB uses an 80/20 split stratified by task type. RG, which defines easy and hard variants of the same 96 generators, is trained on seeds 0--1 and evaluated on seed 2. AppWorld uses the official train and \texttt{test\_normal} splits. BFCL uses category-wise train/test splits for its single- and multi-turn tasks. Full split details are in \cref{sec:appendix_eval} (Appendix \cref{tab:benchmark_details}).

\paragraph{Models.}
In our main results, Qwen3-32B is used as the teacher and Qwen3-4B as the student~\citep{DBLP:journals/corr/abs-2505-09388}. Unless noted otherwise, GRPO is initialised from the 3H SFT checkpoint. \cref{app:gemma} reports corresponding MAB distillation, rehearsal, and GRPO results, together with RG distillation results, using Gemma-3-4B-it~\citep{DBLP:journals/corr/abs-2503-19786} as the student.

\begin{wraptable}[10]{r}{0.54\textwidth}
\vspace{-13pt}
\centering
\footnotesize
\setlength{\tabcolsep}{2pt}
\caption{Training and held-out harnesses for each benchmark.}
\label{tab:harnesses_main}
\begin{tabular}{@{}lll@{}}
\toprule
Benchmark & Training harnesses & Held out \\
\midrule
MedAgentBench & \textsc{parity} / \textsc{fc} / Terminus-2 & \textsc{xml} \\
Reasoning Gym & Terminus-2 / mini-swe / Goose & Aider \\
AppWorld & \textsc{code} / \textsc{fc} / mini-swe & \textsc{xml} \\
BFCL single-turn & \textsc{fc} / \textsc{json} / Terminus-2 & \textsc{xml} \\
BFCL multi-turn & \textsc{fc} / \textsc{json} / \textsc{pycall} & \textsc{xml} \\
\bottomrule
\end{tabular}
\vspace{-8pt}
\end{wraptable}

\paragraph{Harnesses.}
For each benchmark we train on three harnesses and hold out a fourth (\cref{tab:harnesses_main}). Terminus-2~\citep{DBLP:conf/iclr/MerrillSCLRBSSW26}, mini-swe-agent~\citep{DBLP:conf/nips/YangJWLYNP24}, Goose, and Aider are off-the-shelf harnesses supported by Harbor, each with its own system prompts, interaction loop, context management, budgets, and sampling defaults (Appendix \cref{tab:harness_impl}). The remaining harnesses vary mainly in action syntax: \textsc{fc} uses native tool calls, \textsc{json} and \textsc{xml} use structured action formats, \textsc{pycall} uses Python calls, and \textsc{code} uses AppWorld's persistent Python environment. \textsc{parity} is MAB's benchmark-native harness.

\paragraph{Training and evaluation.}
We run single- and multi-harness SFT on all four benchmarks and GRPO on MAB and RG. Harbor supplies the sandboxed task environments, harness implementations, and verifiers for every stage. For SFT, verifier-passing trajectories from each training harness are collected and converted into distillation dataset. For GRPO, the same environments are utilised online through SkyRL's Harbor integration. Each trained model is evaluated through the three training harnesses and the held-out harness on the same environments and verifiers. Our main results report mean success over four attempts ($n=4$) with 95\% task-level bootstrap
confidence intervals (CIs), unless otherwise stated (\cref{sec:appendix_eval}). Each training condition is a single training run unless stated otherwise; the main distillation and rehearsal comparisons are repeated over four independent training runs in \cref{app:seeds}.

\begin{wrapfigure}[20]{r}{0.46\textwidth}
\vspace{-55pt}
\centering
\includegraphics[width=\linewidth]{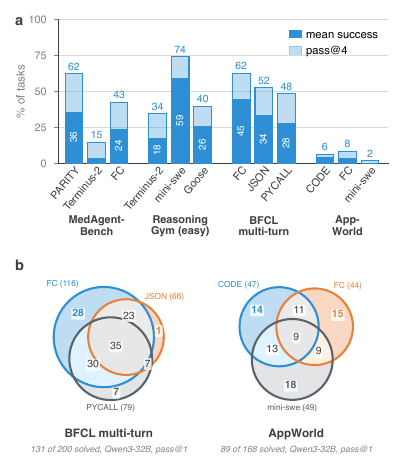}
\caption{\textbf{Harness choice changes task reach.} (a) Qwen3-4B base model performance across harnesses. (b) Tasks each harness solves on datasets, counts per region, Qwen3-32B, pass@1.}
\label{fig:reach}
\end{wrapfigure}

\section{Results}
\label{sec:results}

We address the four questions from \cref{sec:intro} in turn. We first examine how harness choice and harness-specific training shape performance, then use harness rehearsal to probe the task/interface distinction directly, and finally examine how these findings extend to the subsequent RL stage.

\subsection{Harness choice exposes different capabilities of a policy}
\label{sec:res_exposure}

We first examine which tasks the base model can solve before training, asking how harness choice affects overall performance and whether weaker-performing harnesses simply solve subsets of the tasks solved by better-performing harnesses, or instead expose distinct capabilities (Q1).
As \cref{fig:reach}a shows, performance varies substantially for Qwen3-4B across harnesses. On MAB, for example, mean success ranges from about $4\%$ to $36\%$. Importantly, we find that the differences are not merely in overall strength. We find that on every benchmark, weaker-performing harnesses solve tasks that the best-performing harness does not, accounting for 17\% of all solved tasks on MAB, 11\% on BFCL-MT, and 5\% on RG (Appendix \cref{tab:res_reach}). This suggests that harness–policy fit is not one-dimensional. A harness can perform worse overall while still enabling the policy to solve tasks that a stronger harness does not.

\subsection{Distillation Couples Task and Interface Learning}
\label{sec:res_ladder}

\begin{table}[t]
\centering
\small
\setlength{\tabcolsep}{2pt}
\caption{\textbf{Single-harness specialists improve their own interface but often degrade others, while 3H can achieve gains beyond single-harness training across harnesses.}
Rows indicate training conditions and columns evaluation harnesses. Values are mean success rates (\%) with 95\% confidence intervals. Arrows show changes relative to the base model, and bold indicates the best training condition in each column. Results average four training runs with task-bootstrap 95\% CIs. Results for BFCL single-turn and Reasoning Gym Easy are reported in Appendix \cref{tab:res_ladders_app}, and held-out harness results in \cref{fig:res_heldout}.}
\label{tab:res_ladders_main}
\begin{tabular}{@{}lccc @{\hspace{5pt}} lccc@{}}
\toprule
\multicolumn{4}{c}{\textbf{MedAgentBench}} & \multicolumn{4}{c}{\textbf{AppWorld}} \\
\cmidrule(r{6pt}){1-4}\cmidrule(l){5-8}
Training & \textsc{parity} & Terminus-2 & \textsc{fc}
& Training & \textsc{code} & \textsc{fc} & mini-swe \\
\midrule
Base          & 36.1$\pm$9.2 & 3.7$\pm$2.3 & 24.3$\pm$8.4 & Base
& 4.2$\pm$3.9 & 3.9$\pm$3.2 & 0.5$\pm$0.5 \\
\textsc{parity}-only & 45.8$\pm$10.3\up & 7.6$\pm$3.4\up & 18.6$\pm$7.8\dn & \textsc{code}-only
& 12.8$\pm$3.1\up & 9.2$\pm$2.7\up & 7.8$\pm$2.2\up \\
Terminus-2-only & 28.8$\pm$7.7\dn & \textbf{25.3$\pm$7.2}\up & 21.7$\pm$6.8\dn & \textsc{fc}-only
& 2.4$\pm$0.8\dn & 1.7$\pm$0.7\dn & 0.9$\pm$0.5\up \\
\textsc{fc}-only & 28.5$\pm$11.2\dn & 0.0\dn & 48.3$\pm$7.9\up & mini-swe-only
& 14.2$\pm$3.1\up & 7.1$\pm$2.0\up & 12.6$\pm$3.1\up \\
\rowcolor{gray!12}
3H & \textbf{68.0$\pm$10.3}\up & 25.1$\pm$6.6\up & \textbf{52.4$\pm$7.2}\up & 3H
& \textbf{19.5$\pm$3.8}\up & \textbf{16.2$\pm$3.6}\up & \textbf{20.9$\pm$4.1}\up \\
\midrule
\multicolumn{4}{c}{\textbf{BFCL multi-turn}} & \multicolumn{4}{c}{\textbf{Reasoning Gym Hard}} \\
\cmidrule(r{6pt}){1-4}\cmidrule(l){5-8}
Training & \textsc{fc} & \textsc{json} & \textsc{pycall} & Training & Terminus-2 & mini-swe & Goose \\
\midrule
Base & 44.6$\pm$5.9 & 33.5$\pm$5.4 & 28.2$\pm$5.1 & Base & 18.1$\pm$6.1 & 44.1$\pm$7.9 & 20.7$\pm$6.8 \\
\textsc{fc}-only & 53.8$\pm$4.9\up & 0.1$\pm$0.2\dn & 14.2$\pm$3.0\dn & Terminus-2-only & \textbf{40.2$\pm$7.6}\up & 41.5$\pm$7.8\dn & 23.8$\pm$5.6\up \\
\textsc{json}-only & 23.4$\pm$3.9\dn & 47.7$\pm$4.8\up & 30.7$\pm$3.9\up & mini-swe-only & 15.6$\pm$4.2\dn & 35.0$\pm$7.6\dn & 31.7$\pm$7.2\up \\
\textsc{pycall}-only & 31.5$\pm$4.3\dn & 33.8$\pm$4.6\up & 47.7$\pm$4.9\up & Goose-only & 28.0$\pm$6.1\up & 41.1$\pm$7.2\dn & 45.4$\pm$8.2\up \\
\rowcolor{gray!12}
3H & \textbf{60.2$\pm$5.1}\up & \textbf{50.2$\pm$5.2}\up & \textbf{49.1$\pm$4.8}\up & 3H & 39.5$\pm$7.1\up & \textbf{48.1$\pm$8.0}\up & \textbf{45.8$\pm$8.0}\up \\
\bottomrule
\end{tabular}
\end{table}

\begin{wrapfigure}[21]{r}{0.56\textwidth}
\vspace{-17pt}
\centering
\includegraphics[width=\linewidth]{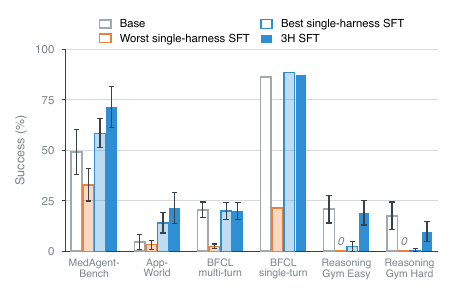}
\caption{\textbf{Transfer to the held-out harness.}
All settings are evaluated on the held-out harness. \emph{worst} and \emph{best} denote those with the lowest and highest held-out scores. The worst specialist falls below half of base performance in four of six settings, while 3H avoids collapse but exceeds base only on MedAgentBench and AppWorld. Full results are in Appendix \cref{tab:res_heldout}.}
\label{fig:res_heldout}
\vspace{-8pt}
\end{wrapfigure}

Q2 asks how harness-conditioned distillation changes performance across interfaces. In a regular setting, each distillation trace jointly supervises the task solution and its execution through a particular harness, so the two effects are coupled under standard training. We first compare single-harness specialists with 3H on the training harnesses (\cref{tab:res_ladders_main}), then evaluate the same policies on held-out harnesses (\cref{fig:res_heldout}).

\textbf{(i) Single-harness distillation produces harness-specific gains.} Across benchmarks, single-harness distillation most often improves performance on the training harness, as in \textsc{parity} on MAB ($36.1\to45.8$) and mini-swe on AppWorld ($0.5\to12.6$). Each trace demonstrates both how to solve the task and how to act through the harness, and performance often improves when evaluation keeps both settings.

\textbf{(ii) But it can impair performance on other harnesses.} Single-harness distillation does not merely fail to transfer. It can actively reduce performance on other harnesses. On BFCL-MT, for example, \textsc{fc} distillation drives \textsc{json} performance from $33.5$ down to $0.1$, well below the base policy. This suggests that specialising to one harness can interfere with competence on others.
\textbf{(iii) Multi-harness distillation can match or exceed specialist training across harnesses.} Compared with training separately on each harness, training a single policy across multiple harnesses could lead to several outcomes: performance may degrade because of interference between harnesses, remain comparable to specialist training, or improve through cross-harness transfer. In our experiments, we observe both comparable performance and gains over specialist training. On BFCL-MT, 3H raises \textsc{json} to $50.2$, above \textsc{json}-only distillation ($47.7$) and well above base ($33.5$). On AppWorld, 3H outperforms each specialist on its own harness (\textsc{code} $19.5$ vs.\ $12.8$, \textsc{fc} $16.2$ vs.\ $1.7$, mini-swe $20.9$ vs.\ $12.6$), and BFCL-MT and RG Hard show similar trends. 
\textbf{(iv) Multi-harness distillation mitigates held-out degradation, with mixed gains beyond the base policy.} We next evaluate on held-out harnesses (\cref{fig:res_heldout}) to ask two questions: whether multi-harness distillation mitigates held-out harness performance degradation seen from specialist training, and whether it improves performance beyond the base policy on them. The first effect is consistent. In four of six settings, the worst specialist falls below half of base performance. On BFCL-MT, for example, \textsc{fc}-only distillation reduces \textsc{xml} success from $20.6$ to $2.5$. In contrast, 3H outperforms the worst specialist by $10$--$66$ points in every setting. Gains over the base policy are less consistent: 3H improves on MAB and AppWorld, stays near base on BFCL, and declines on RG.

These results show that multi-harness distillation reliably reduces cross-harness interference, but does not consistently transfer gains beyond the interfaces seen during training. We examine whether this gap reflects lack of task or interface competence more directly with harness rehearsal in \cref{sec:res_drills}.

\begin{table*}[t]
\centering
\footnotesize
\setlength{\tabcolsep}{5pt}
\renewcommand{\arraystretch}{1.02}

\caption{\textbf{Harness rehearsal enables cross-harness task supervision to transfer.}
Columns report the base model, task supervision with cross-harness distillation (\emph{Cross-task}), target-harness rehearsal (\emph{Rehearsal}), their combination, direct task supervision through the target-harness specialist distillation (\emph{Target-task}), and multi-harness distillation through all three training harnesses (\emph{3H}). Cross-harness task supervision transfers poorly on its own, while rehearsal alone provides limited task learning. Together, they substantially improve target-harness performance. Mean success (\%) with 95\% CIs. Bold marks the best of the three rehearsal-design columns in each row. The target-harness rows averaged over four training runs are in \cref{tab:res_drills_pooled}, and each run individually in \cref{tab:reb_N}.}
\label{tab:res_drills}
\begin{tabular}{@{}lcccccc@{}}
\toprule
Harness
& Base
& Cross-task
& Rehearsal
& \makecell{Cross-task\\+ rehearsal}
& Target-task
& 3H \\
\midrule

\multicolumn{7}{@{}l}{\textbf{MedAgentBench} \quad target: Terminus-2} \\
\textsc{parity}
& 36.1$\pm$9.2
& \textbf{63.5$\pm$10.0}
& 40.4$\pm$10.2
& 52.5$\pm$9.0
& 26.4$\pm$7.6
& 67.2$\pm$10.2 \\
Terminus-2
& 3.7$\pm$2.3
& 2.0$\pm$1.6
& 9.5$\pm$5.7
& \textbf{34.4$\pm$8.2}
& 21.7$\pm$6.4
& 16.9$\pm$5.9 \\
\textsc{fc}
& 24.3$\pm$8.4
& 47.1$\pm$10.0
& 38.5$\pm$10.9
& \textbf{57.0$\pm$9.0}
& 16.1$\pm$5.2
& 55.3$\pm$8.2 \\

\midrule
\multicolumn{7}{@{}l}{\textbf{Reasoning Gym} \quad target: Terminus-2} \\
Terminus-2
& 18.0$\pm$5.5
& 19.8$\pm$5.9
& 20.0$\pm$5.7
& \textbf{37.0$\pm$7.1}
& 54.7$\pm$7.8
& 55.3$\pm$7.9 \\
mini-swe
& 59.4$\pm$7.9
& 58.8$\pm$7.9
& 45.4$\pm$7.3
& \textbf{62.4$\pm$7.8}
& 55.7$\pm$8.4
& 63.0$\pm$7.6 \\
Goose
& 25.7$\pm$6.7
& 58.4$\pm$8.8
& 20.0$\pm$5.8
& \textbf{62.9$\pm$8.5}
& 40.2$\pm$7.3
& 60.8$\pm$8.2 \\

\midrule
\multicolumn{7}{@{}l}{\textbf{BFCL multi-turn} \quad target: \textsc{fc}} \\
\textsc{fc}
& 44.6$\pm$5.9
& 8.5$\pm$2.8
& 32.5$\pm$5.1
& \textbf{58.8$\pm$5.7}
& 55.2$\pm$5.3
& 59.0$\pm$5.5 \\
\textsc{json}
& 33.5$\pm$5.4
& 47.4$\pm$5.8
& 0.0
& \textbf{50.8$\pm$5.9}
& 0.4$\pm$0.6
& 49.5$\pm$5.4 \\
\textsc{pycall}
& 28.2$\pm$5.1
& \textbf{52.5$\pm$5.3}
& 0.1$\pm$0.2
& 50.9$\pm$5.0
& 19.5$\pm$3.9
& 49.1$\pm$5.6 \\

\bottomrule
\end{tabular}
\end{table*}

\subsection{Task and Interface Supervision Interact}
\label{sec:res_drills}
Results from multi-harness distillation leave open why gains transfer inconsistently to held-out harnesses. One possibility is that task knowledge itself fails to transfer. Another is that the policy learns the task but lacks interface competence with the target harness, either due to limited pre-existing interface competence or interference induced by training on other harnesses. Harness rehearsal results support the latter: \emph{Task supervision from other harnesses becomes substantially more effective once the model (re)learns to operate the target interface.} \cref{tab:res_drills} shows this pattern across benchmarks: neither cross-harness task demonstrations nor rehearsal alone substantially outperforms the base policy, but combining task and interface supervision yields much larger gains on the target harness. These results highlight three key observations:

\textbf{(i) Successful task demonstrations need not come from the target harness.} On BFCL-MT, cross-harness distillation reduces target \textsc{fc} performance from $44.6$ to $8.5$ and rehearsal alone yields $32.5$, but combining them raises performance to $58.8$. The same pattern holds for MAB and RG, where the combination reaches $34.4$ and $37.0$, respectively, and it holds across multiple training runs (\cref{app:seeds}).
\textbf{(ii) Rehearsal reproduces the cross-harness interference induced by specialist training.} On BFCL-MT, rehearsing exclusively on \textsc{fc} collapses performance on the remaining harnesses: \textsc{json} drops from $33.5$ to $0.0$, and \textsc{pycall} drops from $28.2$ to $0.1$. This mirrors the degradation observed under the \textsc{fc} specialist, which scores $0.4$ on \textsc{json} and $19.5$ on \textsc{pycall}.

\textbf{(iii) The interaction does not rely on task content in the rehearsal traces.} On RG, rehearsal consists only of transcribing a provided answer, and the rehearsal traces provide no task knowledge that can transfer to evaluation. The same pattern still holds: rehearsal and cross-harness supervision reach $20.0$ and $19.8$ on their own, but $37.0$ when combined. The effect is also not explained by training volume alone (Appendix \cref{tab:reb_E}). On MAB, replacing the target-harness trajectories in 3H with an equal number of rehearsal trajectories raises target performance from $16.9$ to $34.4$. The 2$\times$2 design restricts additional task supervision to cross-harness sources, providing the cleanest test of whether task knowledge transfers once the target interface has been learned. This does not imply that the base student has no relevant task knowledge. When rehearsal alone already recovers most of the attainable performance, as on MedAgentBench with \textsc{fc} as the target and on Reasoning Gym with the 32B rehearsal demonstrator, the potential benefit of additional cross-harness task supervision is naturally more limited (\cref{app:rebuttal_controls}).

These results suggest that the weak transfer in \cref{sec:res_ladder} reflects interference in interface competence rather than loss of task knowledge: cross-harness knowledge becomes actionable once the model learns the target interface. Consistent with this interpretation, the gains on MAB are, on average, comparable on task templates not seen during rehearsal, while the failures eliminated by rehearsal are predominantly interface-related (\cref{app:leakage,app:failure_types}). Thus, task and interface supervision provide distinct but complementary benefits. Example rehearsal instructions and an additional held-out-harness evaluation are provided in \cref{app:drill_examples}.

\subsection{Multi-Harness Competence Persists Through Subsequent RL}
\label{sec:res_rl}

\begin{table}[t]
\centering\footnotesize\setlength{\tabcolsep}{4pt}

\caption{\textbf{Single- and multi-harness GRPO from the 3H-SFT checkpoint and from the base model.} Success (\%) on MedAgentBench and Reasoning Gym (easy / hard). Each GRPO row is the mean of two training seeds unless marked, with 4 attempts per task and matched rollout counts across rows. \textsc{xml} and Aider are held out. $\pm$ denotes a 95\% task-bootstrap CI. CIs are omitted for Reasoning Gym for brevity and range from $\pm$4 to $\pm$8.}
\label{tab:res_singlerl}
\par\medskip\noindent\emph{MedAgentBench}\par\smallskip
\begin{tabular}{@{}lcccc@{}}
\toprule
 & \textsc{parity} & Terminus-2 & \textsc{fc} & \textsc{xml} (held out) \\
\midrule
Base model & 36.1 $\pm$ 9.2 & 5.0 $\pm$ 2.8 & 24.3 $\pm$ 8.4 & 50.1 $\pm$ 11.0 \\
+ parity-only GRPO & 59.8 $\pm$ 10.9 & 5.0 $\pm$ 3.2 & 25.1 $\pm$ 8.1 & 57.2 $\pm$ 11.5 \\
3H SFT & 68.3 $\pm$ 10.1 & 27.3 $\pm$ 7.2 & 56.4 $\pm$ 7.2 & 71.8 $\pm$ 9.1 \\
+ parity-only GRPO & 74.7 $\pm$ 9.2 & 31.9 $\pm$ 7.6 & 61.4 $\pm$ 6.7 & 72.5 $\pm$ 9.1 \\
+ Terminus-2-only GRPO & 70.9 $\pm$ 9.6 & 53.1 $\pm$ 10.4 & 66.9 $\pm$ 7.2 & 69.5 $\pm$ 9.6 \\
+ 3H GRPO & 77.6 $\pm$ 8.4 & 50.8 $\pm$ 9.7 & 85.7 $\pm$ 6.2 & 78.0 $\pm$ 8.1 \\
\bottomrule
\end{tabular}\par
\par\medskip\noindent\emph{Reasoning Gym, easy / hard}\par\smallskip
\begin{tabular}{@{}lcccc@{}}
\toprule
 & Terminus-2 & mini-swe & Goose & Aider (held out) \\
\midrule
Base model & 18.8 / 16.4 & 58.7 / 44.1 & 24.6 / 20.7 & 20.8 / 17.6 \\
+ Terminus-2-only GRPO & 41.1 / 32.0 & 59.9 / 47.9 & 28.7 / 21.7 & 22.0 / 20.7 \\
3H SFT & 55.3 / 43.3 & 63.0 / 51.0 & 60.8 / 46.0 & 19.0 / 9.8 \\
+ Terminus-2-only GRPO & 59.3 / 45.3 & 66.3 / 51.1 & 63.2 / 43.8 & 18.4 / 11.8 \\
+ mini-swe-only GRPO (one seed) & 56.9 / 41.5 & 66.0 / 52.3 & 63.3 / 45.4 & 17.5 / 6.3 \\
+ 3H GRPO & 56.9 / 43.0 & 64.1 / 50.7 & 64.7 / 45.4 & 18.3 / 11.4 \\
\bottomrule
\end{tabular}\par

\end{table}

Our RL experiments ask whether the cross-harness patterns observed in distillation persist under RL, and in particular whether single-harness RL exhibits the same interference pattern as single-harness SFT. \cref{tab:res_singlerl} reports single- and multi-harness GRPO from the 3H-SFT checkpoint, together with single-harness GRPO from the base model, using matched rollout counts.

\textbf{(i) Multi-harness competence learned during distillation carries through RL.} Starting from the 3H-SFT checkpoint, single-harness GRPO preserves performance on the other harnesses, including those not included in GRPO rollouts. We observe no significant degradation on either harnesses seen during SFT or the held-out harness. In several cases, training on one harness also improves performance on harnesses encountered only during SFT. On MAB, for example, \textsc{fc} rises from $56.4$ after 3H SFT to $61.4$ with \textsc{parity} rollouts and to $66.9$ with Terminus-2 rollouts. Multi-harness GRPO is strongest on MAB, giving the best results on \textsc{parity}, \textsc{fc} and held-out \textsc{xml} and staying within three points of the Terminus-2 specialist on Terminus-2, while single- and multi-harness GRPO are comparable on RG. Together, these results show that the multi-harness competence acquired during distillation is retained through subsequent RL.

\textbf{(ii) On-policy RL is more robust to single-harness training than distillation.} The results above leave open how much of this robustness comes from the RL setup itself versus the prior multi-harness SFT initialisation. GRPO results from the base model provides clarification. Unlike single-harness SFT, which can push performance on other harnesses below the base model, single-harness GRPO through the same harness leaves performance on the remaining harnesses at or above the base-model level. More generally, across both the base-model and 3H-SFT initialisations, we do not observe the cross-harness interference pattern seen in the single-harness SFT experiments. This suggests that GRPO is less prone to cross-harness interference under single-harness training, allowing multi-harness interface competence acquired during distillation to carry through subsequent single-harness RL, including on harnesses seen only during distillation and on held-out harnesses.

\section{Conclusion} \label{sec:conclusion}

As harnesses become an increasingly important part of agentic training \citep{DBLP:journals/corr/abs-2609-13739,DBLP:journals/corr/abs-2609-04518,DBLP:journals/corr/abs-2609-09134}, understanding what policies learn through them becomes increasingly important. In this study, we investigate how training through one or multiple harnesses shapes agentic competence, and formulate a two-component view in which task competence can transfer across harnesses, while its expression depends on competence with the target interface. Harness rehearsal provides the clearest evidence for this decomposition. Task supervision collected through other harnesses becomes substantially more useful on a target harness once the model has learned or re-learned how to operate the target harness through the rehearsal: on BFCL-MT (Qwen3-4B), cross-harness task supervision and rehearsal achieve success rates of $8.5$ and $32.5$ on their own but $58.8$ together, and the same pattern holds on MAB ($2.0$ and $9.5$ to $34.4$) and on RG ($19.8$ and $20.0$ to $37.0$). This view also offers a natural explanation for the uneven held-out performance we observe after both single- and multi-harness distillation, as improvements in task competence do not necessarily translate into better performance when training also changes how effectively the policy operates through the evaluation harness: a single-harness specialist can impair performance on another harness from $33.5$ to $0.1$ (BFCL-MT), whereas multi-harness distillation largely avoids this collapse. The interface competence learned at the distillation stage also persists through subsequent GRPO, where single-harness GRPO no longer degrades performance on other harnesses and multi-harness GRPO can sometimes improve performance on the held-out harness.

Overall, our analyses demonstrate how agentic post-training jointly provides task and interface supervision, and show that understanding the decomposition of these two competencies is important for explaining training outcomes when policies are evaluated across multiple harnesses. Task competence can transfer aggregate across harnesses. While interface competence is more sensitive to interference, it can often be acquired independently at substantially lower cost than retraining on the full task set through the target harness. Our study provides insights on how harnesses shape agentic competence during training and motivates a more principled approach to harness use in post-training.

\section*{AI use statement}
In this work, we used generative AI tools to edit and condense text, to design figure layouts and generate plotting code, and to check numerical consistency between the text and the tables. We did not use generative AI to design the experiments, run training or evaluation, or produce the reported results. All AI-assisted text, figures, and code were reviewed and verified by the authors. We take responsibility for the final content of this work, including text, claims, and artefacts produced with the aid of generative AI.

\section*{Reproducibility statement}
\cref{sec:framework,sec:exp_settings} describe the experimental framework, benchmarks, training setup, and evaluation protocol. The appendices provide implementation details, hyperparameters, rehearsal construction, and complete result tables. To ensure reproducibility, code and supporting artefacts are released at \textcolor{blue}{\url{https://github.com/chaeeunlee-io/harness-competence}}.

\section*{Acknowledgements}
C.L.~and A.P.G.~were supported by the United Kingdom Research and Innovation (grant EP/S02431X/1), UKRI Centre for Doctoral Training in Biomedical AI at the University of Edinburgh, School of Informatics.
H.A.R.~was supported by the Engineering and Physical Sciences Research Council (grant EP/S021566/1), UKRI Centre for Doctoral Training in Foundational AI at University College London, Centre for Artificial Intelligence.
P.M. was supported by the Engineering and Physical Sciences Research Council (EPSRC) through the AI Hub in Generative Models (grant number EP/Y028805/1).
We thank Cameron Wheeler at the University of Edinburgh for his helpful input on this work.
For the purpose of open access, the author has applied a creative commons attribution (CC BY) licence to any author accepted manuscript version arising.

\bibliography{iclr2027_conference}

@article{doi:10.1056/AIdbp2500144,
author = {Yixing Jiang  and Kameron C. Black  and Gloria Geng  and Danny Park  and James Zou  and Andrew Y. Ng  and Jonathan H. Chen },
title = {MedAgentBench: A Virtual EHR Environment to Benchmark Medical LLM Agents},
journal = {NEJM AI},
volume = {2},
number = {9},
pages = {AIdbp2500144},
year = {2025},
doi = {10.1056/AIdbp2500144},

URL = {https://ai.nejm.org/doi/full/10.1056/AIdbp2500144},
eprint = {https://ai.nejm.org/doi/pdf/10.1056/AIdbp2500144}
}

@article{team2026harbor,
  title={Harbor: A framework for evaluating and optimizing agents and models in container environments},
  author={{Harbor Framework Team}},
  journal={Zenodo},
  year={2026}
}

@article{DBLP:journals/corr/abs-2605-22166,
  author       = {Tianshi Xu and
                  Huifeng Wen and
                  Meng Li},
  title        = {Adapting the Interface, Not the Model: Runtime Harness Adaptation
                  for Deterministic {LLM} Agents},
  journal      = {CoRR},
  volume       = {abs/2605.22166},
  year         = {2026},
  url          = {https://doi.org/10.48550/arXiv.2605.22166},
  doi          = {10.48550/ARXIV.2605.22166},
  eprinttype   = {arXiv},
  eprint       = {2605.22166},
  bibsource    = {dblp computer science bibliography, https://dblp.org}
}

@article{DBLP:journals/corr/abs-2603-13966,
  author       = {Suhwan Choi and
                  Yunsung Lee and
                  Yubeen Park and
                  Chris Dongjoo Kim and
                  Ranjay Krishna and
                  Dieter Fox and
                  Youngjae Yu},
  title        = {vla-eval: {A} Unified Evaluation Harness for Vision-Language-Action
                  Models},
  journal      = {CoRR},
  volume       = {abs/2603.13966},
  year         = {2026},
  url          = {https://doi.org/10.48550/arXiv.2603.13966},
  doi          = {10.48550/ARXIV.2603.13966},
  eprinttype   = {arXiv},
  eprint       = {2603.13966},
  bibsource    = {dblp computer science bibliography, https://dblp.org}
}

@article{DBLP:journals/corr/abs-2605-08741,
  author       = {Zhengyang Zhao and
                  Lu Ma and
                  Wentao Zhang},
  title        = {Training with Harnesses: On-Policy Harness Self-Distillation for Complex
                  Reasoning},
  journal      = {CoRR},
  volume       = {abs/2605.08741},
  year         = {2026},
  url          = {https://doi.org/10.48550/arXiv.2605.08741},
  doi          = {10.48550/ARXIV.2605.08741},
  eprinttype   = {arXiv},
  eprint       = {2605.08741},
  bibsource    = {dblp computer science bibliography, https://dblp.org}
}

@inproceedings{DBLP:conf/iclr/YaoZYDSN023,
  author       = {Shunyu Yao and
                  Jeffrey Zhao and
                  Dian Yu and
                  Nan Du and
                  Izhak Shafran and
                  Karthik R. Narasimhan and
                  Yuan Cao},
  title        = {{ReAct}: Synergizing Reasoning and Acting in Language Models},
  booktitle    = {The Eleventh International Conference on Learning Representations,
                  {ICLR} 2023, Kigali, Rwanda, May 1-5, 2023},
  publisher    = {OpenReview.net},
  year         = {2023},
  url          = {https://openreview.net/forum?id=WE\_vluYUL-X},
  bibsource    = {dblp computer science bibliography, https://dblp.org}
}

@inproceedings{DBLP:conf/nips/SchickDDRLHZCS23,
  author       = {Timo Schick and
                  Jane Dwivedi{-}Yu and
                  Roberto Dess{\`{\i}} and
                  Roberta Raileanu and
                  Maria Lomeli and
                  Eric Hambro and
                  Luke Zettlemoyer and
                  Nicola Cancedda and
                  Thomas Scialom},
  editor       = {Alice Oh and
                  Tristan Naumann and
                  Amir Globerson and
                  Kate Saenko and
                  Moritz Hardt and
                  Sergey Levine},
  title        = {Toolformer: Language Models Can Teach Themselves to Use Tools},
  booktitle    = {Advances in Neural Information Processing Systems 36: Annual Conference
                  on Neural Information Processing Systems 2023, NeurIPS 2023, New Orleans,
                  LA, USA, December 10 - 16, 2023},
  year         = {2023},
  url          = {http://papers.nips.cc/paper\_files/paper/2023/hash/d842425e4bf79ba039352da0f658a906-Abstract-Conference.html},
  bibsource    = {dblp computer science bibliography, https://dblp.org}
}

@inproceedings{DBLP:conf/icml/WangCY0L0J24,
  author       = {Xingyao Wang and
                  Yangyi Chen and
                  Lifan Yuan and
                  Yizhe Zhang and
                  Yunzhu Li and
                  Hao Peng and
                  Heng Ji},
  editor       = {Ruslan Salakhutdinov and
                  Zico Kolter and
                  Katherine A. Heller and
                  Adrian Weller and
                  Nuria Oliver and
                  Jonathan Scarlett and
                  Felix Berkenkamp},
  title        = {Executable Code Actions Elicit Better {LLM} Agents},
  booktitle    = {Forty-first International Conference on Machine Learning, {ICML} 2024,
                  Vienna, Austria, July 21-27, 2024},
  series       = {Proceedings of Machine Learning Research},
  volume       = {235},
  pages        = {50208--50232},
  publisher    = {{PMLR} / OpenReview.net},
  year         = {2024},
  url          = {https://proceedings.mlr.press/v235/wang24h.html},
  bibsource    = {dblp computer science bibliography, https://dblp.org}
}

@inproceedings{DBLP:conf/nips/YangJWLYNP24,
  author       = {John Yang and
                  Carlos E. Jimenez and
                  Alexander Wettig and
                  Kilian Lieret and
                  Shunyu Yao and
                  Karthik Narasimhan and
                  Ofir Press},
  editor       = {Amir Globerson and
                  Lester Mackey and
                  Danielle Belgrave and
                  Angela Fan and
                  Ulrich Paquet and
                  Jakub M. Tomczak and
                  Cheng Zhang},
  title        = {{SWE-agent}: Agent-Computer Interfaces Enable Automated Software Engineering},
  booktitle    = {Advances in Neural Information Processing Systems 37: Annual Conference
                  on Neural Information Processing Systems 2024, NeurIPS 2024, Vancouver,
                  BC, Canada, December 10 - 15, 2024},
  year         = {2024},
  url          = {http://papers.nips.cc/paper\_files/paper/2024/hash/5a7c947568c1b1328ccc5230172e1e7c-Abstract-Conference.html},
  bibsource    = {dblp computer science bibliography, https://dblp.org}
}

@inproceedings{DBLP:conf/iclr/0001LSXTZPSLSTL25,
  author       = {Xingyao Wang and
                  Boxuan Li and
                  Yufan Song and
                  Frank F. Xu and
                  Xiangru Tang and
                  Mingchen Zhuge and
                  Jiayi Pan and
                  Yueqi Song and
                  Bowen Li and
                  Jaskirat Singh and
                  Hoang H. Tran and
                  Fuqiang Li and
                  Ren Ma and
                  Mingzhang Zheng and
                  Bill Qian and
                  Yanjun Shao and
                  Niklas Muennighoff and
                  Yizhe Zhang and
                  Binyuan Hui and
                  Junyang Lin and
                  et al.},
  title        = {{OpenHands}: An Open Platform for {AI} Software Developers as Generalist
                  Agents},
  booktitle    = {The Thirteenth International Conference on Learning Representations,
                  {ICLR} 2025, Singapore, April 24-28, 2025},
  publisher    = {OpenReview.net},
  year         = {2025},
  url          = {https://openreview.net/forum?id=OJd3ayDDoF},
  bibsource    = {dblp computer science bibliography, https://dblp.org}
}

@inproceedings{DBLP:conf/iclr/MerrillSCLRBSSW26,
  author       = {Mike A. Merrill and
                  Alexander Glenn Shaw and
                  Nicholas Carlini and
                  Boxuan Li and
                  Harsh Raj and
                  Ivan Bercovich and
                  Lin Shi and
                  Jeong Yeon Shin and
                  Thomas Walshe and
                  Estefany Kelly Buchanan and
                  Junhong Shen and
                  Guanghao Ye and
                  Haowei Lin and
                  Jason Poulos and
                  Maoyu Wang and
                  Marianna Nezhurina and
                  Di Lu and
                  Orfeas Menis{-}Mastromichalakis and
                  Zhiwei Xu and
                  Zizhao Chen and
                  Yue Liu and
                  Robert Zhang and
                  Leon Liangyu Chen and
                  Anurag Kashyap and
                  Jan{-}Lucas Uslu and
                  Jeffrey Li and
                  Jianbo Wu and
                  Minghao Yan and
                  Song Bian and
                  Vedang Sharma and
                  Ke Sun and
                  Steven Dillmann and
                  Akshay Anand and
                  Andrew Lanpouthakoun and
                  Bardia Koopah and
                  Changran Hu and
                  Etash Kumar Guha and
                  Gabriel H. S. Dreiman and
                  Jiacheng Zhu and
                  Karl Krauth and
                  Li Zhong and
                  Niklas Muennighoff and
                  Robert Amanfu and
                  Shangyin Tan and
                  Shreyas Pimpalgaonkar and
                  Tushar Aggarwal and
                  Xiangning Lin and
                  Xin Lan and
                  Xuandong Zhao and
                  Yiqing Liang and
                  Yuanli Wang and
                  Zilong Wang and
                  Changzhi Zhou and
                  David Heineman and
                  Hange Liu and
                  Harsh Trivedi and
                  John Yang and
                  Junhong Lin and
                  Manish Shetty and
                  Michael Yang and
                  Nabil Omi and
                  Negin Raoof and
                  Shanda Li and
                  Terry Yue Zhuo and
                  Wuwei Lin and
                  Yiwei Dai and
                  Yuxin Wang and
                  Wenhao Chai and
                  Shang Zhou and
                  Dariush Wahdany and
                  Ziyu She and
                  Jiaming Hu and
                  Zhikang Dong and
                  Yuxuan Zhu and
                  Sasha Cui and
                  Ahson Saiyed and
                  Arinbj{\"{o}}rn Kolbeinsson and
                  Christopher Michael Rytting and
                  Ryan Marten and
                  Yixin Wang and
                  Jenia Jitsev and
                  Alex Dimakis and
                  Andy Konwinski and
                  Ludwig Schmidt},
  editor       = {Carl Vondrick and
                  Bharath Hariharan and
                  Colin Raffel and
                  Lerrel Pinto and
                  Diyi Yang and
                  Aleksandra Faust},
  title        = {Terminal-Bench: Benchmarking Agents on Hard, Realistic Tasks in Command
                  Line Interfaces},
  booktitle    = {The Fourteenth International Conference on Learning Representations,
                  {ICLR} 2026, Rio de Janeiro, Brazil, April 23-27, 2026},
  publisher    = {proceedings.iclr.cc},
  year         = {2026},
  url          = {https://proceedings.iclr.cc/paper\_files/paper/2026/hash/444a3737adaee10d86ad2ef5f74468e6-Abstract-Conference.html},
  bibsource    = {dblp computer science bibliography, https://dblp.org}
}

@article{DBLP:journals/corr/abs-2402-03300,
  author       = {Zhihong Shao and
                  Peiyi Wang and
                  Qihao Zhu and
                  Runxin Xu and
                  Junxiao Song and
                  Mingchuan Zhang and
                  Y. K. Li and
                  Y. Wu and
                  Daya Guo},
  title        = {{DeepSeekMath}: Pushing the Limits of Mathematical Reasoning in Open
                  Language Models},
  journal      = {CoRR},
  volume       = {abs/2402.03300},
  year         = {2024},
  url          = {https://doi.org/10.48550/arXiv.2402.03300},
  doi          = {10.48550/ARXIV.2402.03300},
  eprinttype   = {arXiv},
  eprint       = {2402.03300},
  bibsource    = {dblp computer science bibliography, https://dblp.org}
}

@article{DBLP:journals/corr/abs-2511-16108,
  author       = {Shiyi Cao and
                  Dacheng Li and
                  Fangzhou Zhao and
                  Shuo Yuan and
                  Sumanth Hegde and
                  Connor Chen and
                  Charlie Ruan and
                  Tyler Griggs and
                  Shu Liu and
                  Eric Tang and
                  Richard Liaw and
                  Philipp Moritz and
                  Matei Zaharia and
                  Joseph E. Gonzalez and
                  Ion Stoica},
  title        = {{SkyRL-Agent}: Efficient {RL} Training for Multi-turn {LLM} Agent},
  journal      = {CoRR},
  volume       = {abs/2511.16108},
  year         = {2025},
  url          = {https://doi.org/10.48550/arXiv.2511.16108},
  doi          = {10.48550/ARXIV.2511.16108},
  eprinttype   = {arXiv},
  eprint       = {2511.16108},
  bibsource    = {dblp computer science bibliography, https://dblp.org}
}

@article{DBLP:journals/corr/abs-2505-09388,
  author       = {{Qwen Team}},
  title        = {Qwen3 Technical Report},
  journal      = {CoRR},
  volume       = {abs/2505.09388},
  year         = {2025},
  url          = {https://doi.org/10.48550/arXiv.2505.09388},
  doi          = {10.48550/ARXIV.2505.09388},
  eprinttype   = {arXiv},
  eprint       = {2505.09388},
  bibsource    = {dblp computer science bibliography, https://dblp.org}
}

@article{DBLP:journals/corr/abs-2503-19786,
  author       = {{Gemma Team}},
  title        = {Gemma 3 Technical Report},
  journal      = {CoRR},
  volume       = {abs/2503.19786},
  year         = {2025},
  url          = {https://doi.org/10.48550/arXiv.2503.19786},
  doi          = {10.48550/ARXIV.2503.19786},
  eprinttype   = {arXiv},
  eprint       = {2503.19786},
  bibsource    = {dblp computer science bibliography, https://dblp.org}
}

@article{DBLP:journals/corr/abs-2609-13739,
  author       = {Hongliang Wei and
                  Xiaobing Tu and
                  Yinggui Wang and
                  Zhengxi Liu and
                  Rongkun Xue and
                  Jinkui Ren and
                  Xiantao Zhang and
                  Debin Zhao and
                  Xiaopeng Fan},
  title        = {{HarnessBandit}: Joint Learnability-Transferability Scheduling for Multi-Harness
                  Agentic Reinforcement Learning},
  journal      = {CoRR},
  volume       = {abs/2609.13739},
  year         = {2026},
  url          = {https://doi.org/10.48550/arXiv.2609.13739},
  doi          = {10.48550/ARXIV.2609.13739},
  eprinttype   = {arXiv},
  eprint       = {2609.13739},
  bibsource    = {dblp computer science bibliography, https://dblp.org}
}

@article{DBLP:journals/corr/abs-2609-04518,
  author       = {Chenqian Le and
                  Jiayi Cheng and
                  Qijia He and
                  Runhao Li and
                  Yinghao Li and
                  Xupeng Chen},
  title        = {What Does Multi-Harness {RL} Learn? Credit Assignment and Portability
                  in Coding Agents},
  journal      = {CoRR},
  volume       = {abs/2609.04518},
  year         = {2026},
  url          = {https://doi.org/10.48550/arXiv.2609.04518},
  doi          = {10.48550/ARXIV.2609.04518},
  eprinttype   = {arXiv},
  eprint       = {2609.04518},
  bibsource    = {dblp computer science bibliography, https://dblp.org}
}

@article{DBLP:journals/corr/abs-2606-14249,
  author       = {Tingyang Chen and
                  Shuo Lu and
                  Kang Zhao and
                  Weicheng Meng and
                  Hanlin Teng and
                  Tianhao Li and
                  Chao Li and
                  Xule Liu and
                  Jian Liang and
                  Zhizhong Zhang and
                  Yuan Xie and
                  Heng Qu and
                  Kun Shao and
                  Jian Luan},
  title        = {{HarnessX}: {A} Composable, Adaptive, and Evolvable Agent Harness Foundry},
  journal      = {CoRR},
  volume       = {abs/2606.14249},
  year         = {2026},
  url          = {https://doi.org/10.48550/arXiv.2606.14249},
  doi          = {10.48550/ARXIV.2606.14249},
  eprinttype   = {arXiv},
  eprint       = {2606.14249},
  bibsource    = {dblp computer science bibliography, https://dblp.org}
}

@article{DBLP:journals/corr/abs-2609-09134,
  author       = {Zhou Yu and
                  Bin Bi and
                  Shiva Kumar Pentyala and
                  Shubham Mehrotra and
                  Sougata Chaudhuri and
                  Shilpa Bhagavath and
                  Zeyuan Chen and
                  Ran Xu and
                  Phil Mui and
                  James Zhu and
                  Sitaram Asur},
  title        = {Co-Evolving Harnesses and Models: On-Policy Correction Helps Weaker
                  Models Catch Up Where Imitation Fails},
  journal      = {CoRR},
  volume       = {abs/2609.09134},
  year         = {2026},
  url          = {https://doi.org/10.48550/arXiv.2609.09134},
  doi          = {10.48550/ARXIV.2609.09134},
  eprinttype   = {arXiv},
  eprint       = {2609.09134},
  bibsource    = {dblp computer science bibliography, https://dblp.org}
}

@inproceedings{DBLP:conf/icml/PatilMYJSSG25,
  author       = {Shishir G. Patil and
                  Huanzhi Mao and
                  Fanjia Yan and
                  Charlie Cheng{-}Jie Ji and
                  Vishnu Suresh and
                  Ion Stoica and
                  Joseph E. Gonzalez},
  editor       = {Aarti Singh and
                  Maryam Fazel and
                  Daniel Hsu and
                  Simon Lacoste{-}Julien and
                  Felix Berkenkamp and
                  Tegan Maharaj and
                  Kiri Wagstaff and
                  Jerry Zhu},
  title        = {The Berkeley Function Calling Leaderboard {(BFCL):} From Tool Use
                  to Agentic Evaluation of Large Language Models},
  booktitle    = {Forty-second International Conference on Machine Learning, {ICML}
                  2025, Vancouver, BC, Canada, July 13-19, 2025},
  series       = {Proceedings of Machine Learning Research},
  volume       = {267},
  publisher    = {{PMLR} / OpenReview.net},
  year         = {2025},
  url          = {https://proceedings.mlr.press/v267/patil25a.html},
  bibsource    = {dblp computer science bibliography, https://dblp.org}
}

@inproceedings{DBLP:conf/acl/TrivediKHMDLGSB24,
  author       = {Harsh Trivedi and
                  Tushar Khot and
                  Mareike Hartmann and
                  Ruskin Manku and
                  Vinty Dong and
                  Edward Li and
                  Shashank Gupta and
                  Ashish Sabharwal and
                  Niranjan Balasubramanian},
  editor       = {Lun{-}Wei Ku and
                  Andre Martins and
                  Vivek Srikumar},
  title        = {{AppWorld}: {A} Controllable World of Apps and People for Benchmarking
                  Interactive Coding Agents},
  booktitle    = {Proceedings of the 62nd Annual Meeting of the Association for Computational
                  Linguistics (Volume 1: Long Papers), {ACL} 2024, Bangkok, Thailand,
                  August 11-16, 2024},
  pages        = {16022--16076},
  publisher    = {Association for Computational Linguistics},
  year         = {2024},
  url          = {https://doi.org/10.18653/v1/2024.acl-long.850},
  doi          = {10.18653/V1/2024.ACL-LONG.850},
  bibsource    = {dblp computer science bibliography, https://dblp.org}
}

@inproceedings{DBLP:conf/nips/StojanovskiSSJA25,
  author       = {Zafir Stojanovski and
                  Oliver Stanley and
                  Joe Sharratt and
                  Richard Jones and
                  Abdulhakeem Adefioye and
                  Jean Kaddour and
                  Andreas K{\"{o}}pf},
  editor       = {Danielle Belgrave and
                  Cheng Zhang and
                  Laura N. Montoya and
                  Hsuan{-}Tien Lin and
                  Razvan Pascanu and
                  Piotr Koniusz and
                  Marzyeh Ghassemi and
                  Nancy Chen and
                  Iv{\'{a}}n Vladimir Meza Ru{\'{\i}}z and
                  Arturo Loaiza{-}Bonilla},
  title        = {Reasoning Gym: Reasoning Environments for Reinforcement Learning with
                  Verifiable Rewards},
  booktitle    = {Advances in Neural Information Processing Systems 38: Annual Conference
                  on Neural Information Processing Systems 2025, NeurIPS 2025, San Diego,
                  CA, USA, December 2-7, 2025 / Mexico City, Mexico, November 30 - December
                  5, 2025},
  year         = {2025},
  url          = {http://papers.nips.cc/paper\_files/paper/2025/hash/528d56195a2c77c808494c86fa7c77ad-Abstract-Datasets\_and\_Benchmarks\_Track.html},
  bibsource    = {dblp computer science bibliography, https://dblp.org}
}

@article{DBLP:journals/corr/abs-2605-27922,
  author       = {Yilun Yao and
                  Xinyu Tan and
                  Chao{-}Hsuan Liu and
                  Yaoming Li and
                  Zhengyang Wang and
                  Wenhan Yu and
                  Zhewen Tan and
                  Yuxuan Tian and
                  Guangxiang Zhao and
                  Lin Sun and
                  Xiangzheng Zhang and
                  Tong Yang},
  title        = {Harness-Bench: Measuring Harness Effects across Models in Realistic
                  Agent Workflows},
  journal      = {CoRR},
  volume       = {abs/2605.27922},
  year         = {2026},
  url          = {https://doi.org/10.48550/arXiv.2605.27922},
  doi          = {10.48550/ARXIV.2605.27922},
  eprinttype   = {arXiv},
  eprint       = {2605.27922},
  bibsource    = {dblp computer science bibliography, https://dblp.org}
}

@article{DBLP:journals/corr/abs-2609-01437,
  author       = {Yuhao Wu and
                  Jingyuan Zhang and
                  Jiajun Shi and
                  Xinping Lei and
                  Qingshui Gu and
                  Yuxuan Zhang and
                  Zexuan Wang and
                  Chen He and
                  Chen Huang and
                  Maojia Song and
                  Zhiyuan Zeng and
                  Shaowen Wang and
                  Jinkai Liu and
                  Yunfeng Shi and
                  Jiaheng Liu and
                  Shen Yan and
                  Wenhao Huang and
                  Ge Zhang and
                  Wenxuan Zhang},
  title        = {{HarnessDev}: Can {LLMs} Create and Evolve Their Own Agent Harness?},
  journal      = {CoRR},
  volume       = {abs/2609.01437},
  year         = {2026},
  url          = {https://doi.org/10.48550/arXiv.2609.01437},
  doi          = {10.48550/ARXIV.2609.01437},
  eprinttype   = {arXiv},
  eprint       = {2609.01437},
  bibsource    = {dblp computer science bibliography, https://dblp.org}
}

@article{DBLP:journals/corr/abs-2603-28052,
  author       = {Yoonho Lee and
                  Roshen Nair and
                  Qizheng Zhang and
                  Kangwook Lee and
                  Omar Khattab and
                  Chelsea Finn},
  title        = {Meta-Harness: End-to-End Optimization of Model Harnesses},
  journal      = {CoRR},
  volume       = {abs/2603.28052},
  year         = {2026},
  url          = {https://doi.org/10.48550/arXiv.2603.28052},
  doi          = {10.48550/ARXIV.2603.28052},
  eprinttype   = {arXiv},
  eprint       = {2603.28052},
  bibsource    = {dblp computer science bibliography, https://dblp.org}
}

@article{DBLP:journals/corr/abs-2606-09498,
  author       = {Hangfan Zhang and
                  Shao Zhang and
                  Kangcong Li and
                  Chen Zhang and
                  Yang Chen and
                  Yiqun Zhang and
                  Lei Bai and
                  Shuyue Hu},
  title        = {Self-Harness: Harnesses That Improve Themselves},
  journal      = {CoRR},
  volume       = {abs/2606.09498},
  year         = {2026},
  url          = {https://doi.org/10.48550/arXiv.2606.09498},
  doi          = {10.48550/ARXIV.2606.09498},
  eprinttype   = {arXiv},
  eprint       = {2606.09498},
  bibsource    = {dblp computer science bibliography, https://dblp.org}
}

@article{DBLP:journals/corr/abs-2608-06301,
  author       = {Varun Ursekar and
                  Apaar Shanker and
                  Yash Maurya and
                  Shehab Yasser and
                  Vijay S. Kalmath and
                  Veronica Chatrath and
                  Yuan Xue},
  title        = {{HarnessOpt-Bench}: Evaluating {LLMs} at Harness Optimization},
  journal      = {CoRR},
  volume       = {abs/2608.06301},
  year         = {2026},
  url          = {https://doi.org/10.48550/arXiv.2608.06301},
  doi          = {10.48550/ARXIV.2608.06301},
  eprinttype   = {arXiv},
  eprint       = {2608.06301},
  bibsource    = {dblp computer science bibliography, https://dblp.org}
}

@article{DBLP:journals/corr/abs-2607-28802,
  author       = {Harsh Raj and
                  Vipul Gupta and
                  Anas Mahmoud and
                  Razvan{-}Gabriel Dumitru and
                  Darvin Yi and
                  Aakash Sabharwal and
                  Yunzhong He},
  title        = {Model or Harness? An Interaction-Centric Taxonomy for Localizing Agent
                  Failures},
  journal      = {CoRR},
  volume       = {abs/2607.28802},
  year         = {2026},
  url          = {https://doi.org/10.48550/arXiv.2607.28802},
  doi          = {10.48550/ARXIV.2607.28802},
  eprinttype   = {arXiv},
  eprint       = {2607.28802},
  bibsource    = {dblp computer science bibliography, https://dblp.org}
}

@article{DBLP:journals/corr/abs-2503-20783,
  author       = {Zichen Liu and
                  Changyu Chen and
                  Wenjun Li and
                  Penghui Qi and
                  Tianyu Pang and
                  Chao Du and
                  Wee Sun Lee and
                  Min Lin},
  title        = {Understanding R1-Zero-Like Training: {A} Critical Perspective},
  journal      = {CoRR},
  volume       = {abs/2503.20783},
  year         = {2025},
  url          = {https://doi.org/10.48550/arXiv.2503.20783},
  doi          = {10.48550/ARXIV.2503.20783},
  eprinttype   = {arXiv},
  eprint       = {2503.20783},
  bibsource    = {dblp computer science bibliography, https://dblp.org}
}
\bibliographystyle{iclr2027_conference}

\appendix

\section{Datasets, Harnesses, and Evaluation}
\label[appendix]{sec:appendix_eval}

\subsection{Datasets and Evaluation}
\label[appendix]{sec:appendix_benchmarks}

\cref{tab:benchmark_details} summarises dataset splits and evaluation protocol for each benchmark.

\Needspace{18\baselineskip}
\begin{wraptable}{r}{0.50\textwidth}
\vspace{-10pt}
\centering
\footnotesize
\setlength{\tabcolsep}{2pt}
\caption{Dataset splits and main evaluation protocols. Results presented in the main text use $n=4$ attempts at $T=0.7$. BFCL single-turn uses greedy decoding ($n=1$, $T=0$).}
\label{tab:benchmark_details}
\begin{tabular}{@{}lrrl@{}}
\toprule
Benchmark & Train & Test & Main evaluation \\
\midrule
MedAgentBench & 209 & 61 & $n=4$, $T=0.7$ \\
Reasoning Gym (per tier) & 192 & 96 & $n=4$, $T=0.7$ \\
AppWorld & 94 & 168 & $n=4$, $T=0.7$ \\
BFCL single-turn & 1,051 & 355 & $n=1$, $T=0$ \\
BFCL multi-turn & 600 & 200 & $n=4$, $T=0.7$ \\
\bottomrule
\end{tabular}

\vspace{-8pt}
\end{wraptable}

\paragraph{Task splits.} MAB uses 209 training and 61 test instances.
The training split contains 215 instances before filtering. RG uses 96
generator families, with seeds 0--1 yielding 192 training instances and
seed 2 yielding 96 test instances per difficulty configuration.
BFCL V3 splits use seed 20260907. For single-turn, capping each of 13 categories
at 150 examples reduces 3,641 examples to 1,406, split into 1,051 training
and 355 test examples. Multi-turn uses 800 examples from four categories,
split into 600 training and 200 test examples. AppWorld uses the unmodified
94-task training and 168-task \texttt{test\_normal} splits.

\paragraph{Confidence intervals.} Unless otherwise noted, we compute 95\% confidence intervals using a nonparametric percentile bootstrap with \(B{=}1{,}000\) replicates and a fixed seed. We resample tasks for all benchmarks except AppWorld. For AppWorld, we instead resample its 56 scenarios because each scenario contributes three task variants that share world state. We report the 2.5th and 97.5th percentiles, with printed \(\pm\) values denoting half the interval width. Exact-zero cells omit intervals, and BFCL-ST is reported without bootstrap intervals.

\subsection{Harness Details}
\label[appendix]{sec:appendix_harnesses}

\cref{tab:harness_impl} summarises harness implementations across all benchmarks.

\paragraph{Harness types.} The harnesses broadly fall into three types. (i) \emph{Off-the-shelf agent frameworks provided through Harbor}: Terminus-2 (version 2.0.0), mini-swe-agent (2.4.6), Goose (1.51.0), and Aider (0.86.2). We use each framework as provided. Each defines its own system prompt, action and observation formats, history handling, and budgets, with the model interacting through a shell or file interface. (ii) \emph{Custom adapters designed to vary only the interaction format}: \textsc{parity}, \textsc{fc}, \textsc{json}, \textsc{pycall}, and \textsc{xml}. Within each benchmark, these adapters use the same task text, function set, environment, verifier, and budgets. They differ only in the system prompt, action syntax, and observation formatting. Task construction, execution, and scoring are otherwise identical. (iii) The AppWorld harnesses \textsc{code}, \textsc{fc} and \textsc{xml}, which wrap the same persistent Python session. \textsc{code} uses a code block, \textsc{fc} uses a single \texttt{execute\_python} tool call, and \textsc{xml} uses tagged code. All three expose the same action space.

All four RG harnesses are off-the-shelf agent frameworks. BFCL-MT uses only custom adapters that vary the interaction format. MAB, BFCL-ST, and AppWorld each combine custom adapters with one off-the-shelf agent framework: Terminus-2 for MAB and BFCL-ST, and mini-swe-agent for AppWorld. \textsc{pycall} is a text-based action format: the adapter parses Python-style function-call syntax and invokes the corresponding functions without running a Python interpreter. This differs from AppWorld \textsc{code}, which executes Python directly.

\begin{table}[t]
\centering
\scriptsize
\setlength{\tabcolsep}{2.5pt}
\caption{\textbf{Harness implementations.} Summary of each harness implementation. MedAgentBench harnesses act against a
live FHIR server inside the trial container and submit a final answer. Reasoning Gym uses off-the-shelf agent frameworks that write final answers to \texttt{/workspace/answer.txt}. BFCL multi-turn uses stateful API instances within each trial. AppWorld \textsc{code}, \textsc{fc}, and \textsc{xml} share the same persistent Python session. Held-out harnesses are marked with $\dagger$.}
\label{tab:harness_impl}
\begin{tabular}{@{}>{\raggedright\arraybackslash}p{0.55in}>{\raggedright\arraybackslash}p{0.85in}>{\raggedright\arraybackslash}p{1.2in}>{\raggedright\arraybackslash}p{0.8in}>{\raggedright\arraybackslash}p{0.7in}>{\raggedright\arraybackslash}p{0.95in}@{}}
\toprule
Harness & Kind & Acts through & State across turns & History & Termination \\
\midrule
\multicolumn{6}{@{}l}{\textit{MedAgentBench}} \\
\textsc{parity} & custom adapter & text: \texttt{GET}, \texttt{POST}, \texttt{FINISH(...)} & server state only & full, resent each turn & \texttt{FINISH}, 8 rounds, or a third unparseable reply \\
\textsc{fc} & custom adapter & native tool calls & server state only & full & \texttt{finish} call or 8 rounds \\
Terminus-2 & released (2.0.0) & terminal: JSON commands run in a shell & shell and filesystem & agent-managed, summarised & agent marks complete, or budget ends \\
\textsc{xml}$^\dagger$ & custom adapter & XML tags & server state only & full & \texttt{<finish>}, 8 rounds, or a third unparseable reply \\
\midrule
\multicolumn{6}{@{}l}{\textit{Reasoning Gym}} \\
Terminus-2 & released (2.0.0) & terminal: JSON commands run in a shell & shell and filesystem & agent-managed, summarised & agent stops or budget ends \\
mini-swe & released (2.4.6) & native tool calls to a shell tool & shell and filesystem & agent-managed & agent stops or budget ends \\
Goose & released (1.51.0) & native tool calls, including shell and file editing & session and filesystem & agent-managed & agent stops or budget ends \\
Aider$^\dagger$ & released (0.86.2) & the agent's edit and command interface & filesystem & agent-managed & agent stops or budget ends \\
\midrule
\multicolumn{6}{@{}l}{\textit{BFCL single-turn}} \\
\textsc{fc} & custom adapter & native tool calls & none & one exchange & after one response \\
\textsc{json} & custom adapter & JSON object with a list of calls & none & one exchange & after one response \\
Terminus-2 & released (2.0.0) & terminal: JSON commands run in a shell & shell and filesystem & agent-managed, summarised & completion, or 8 episodes \\
\textsc{xml}$^\dagger$ & custom adapter & XML-tagged calls & none & one exchange & after one response \\
\midrule
\multicolumn{6}{@{}l}{\textit{BFCL multi-turn}} \\
\textsc{fc} & custom adapter & native tool calls & stateful API instances & full & reply with no calls ends the user turn, 20 steps per turn, episode ends after the last user turn \\
\textsc{json} & custom adapter & \texttt{\{"calls": [...]\}} & stateful API instances & full & same \\
\textsc{pycall} & custom adapter & python-style call lines in a fenced block (parsed, not executed) & stateful API instances & full & same \\
\textsc{xml}$^\dagger$ & custom adapter & \texttt{<calls>}/\texttt{<call>} tags & stateful API instances & full & same \\
\midrule
\multicolumn{6}{@{}l}{\textit{AppWorld}} \\
\textsc{code} & custom adapter & Python in a fenced block, run in an IPython session & session and world state & full & \texttt{complete\_task}, 50 steps, or 40 errors \\
\textsc{fc} & custom adapter & one \texttt{execute\_python} tool call, same session & session and world state & full & same \\
mini-swe & adapted from release (2.4.6) & Python in a fenced block, sent to the same session & session and world state & framework's own & framework's own \\
\textsc{xml}$^\dagger$ & custom adapter & \texttt{<reasoning>}/\texttt{<code>} tags, same session & session and world state & full & same \\
\bottomrule
\end{tabular}
\end{table}

\section{Training Details}
\label[appendix]{app:training}

This section provides additional details on the SFT and RL experiments, including training configurations, data collection, and the experimental frameworks summarised in \cref{sec:exp_settings}.

\paragraph{Framework.} Each task is implemented as a containerised Harbor directory with a declarative configuration, a natural-language instruction, an environment image, and a verifier that returns a binary reward. SFT trajectory collection, RL rollouts, and evaluation use the same task directories, harness implementations, and verifiers. Rehearsal uses the same setup, with only the task instruction replaced (\cref{app:drill_examples}). For RL, SkyRL provides the trainer and Harbor executes the rollouts. Agents query the current policy through a colocated inference engine, with updated weights propagated between training steps.

\subsection{Supervised Fine-Tuning}
\label[appendix]{sec:appendix_training}

\cref{tab:sft_hparams} summarises the SFT setup and hyperparameters used across all distillation experiments and benchmarks. \cref{tab:pool_inventory} summarises the distillation datasets and the collection settings used to construct them.

\begin{table}[t]
\centering
\small
\caption{Supervised fine-tuning setup used across all distillation experiments.}
\label{tab:sft_hparams}
\begin{tabular}{@{}ll@{}}
\toprule
Setting & Value \\
\midrule
epochs            & 2 \\
learning rate     & $1\times10^{-5}$ \\
LR schedule       & cosine, 10 warmup steps \\
batch / grad.\ accum. & $1$ / $8$ \\
sequence length   & $32{,}768$ tokens \\
optimiser         & paged AdamW (8-bit) \\
kernels           & Liger \\
precision         & bf16 \\
hardware          & $1\times$ H100 (80\,GB) \\
chat template     & native, tool-call for Gemma \textsc{fc} \\
\bottomrule
\end{tabular}
\end{table}

\begin{table}[t]
\centering\small
\caption{Distillation data by benchmark, reported as retained trajectories / distinct training tasks. Columns (a)--(c) are the three single-harness datasets in the order given in the second column (e.g.\ for MedAgentBench: (a)~\textsc{parity}, (b)~Terminus-2, and (c)~\textsc{fc}). The final column combines all three harnesses (3H), with task counts taken over the union of (a)--(c).}
\label{tab:pool_inventory}
\begin{tabular}{@{}llrrrr@{}}
\toprule
Benchmark & Harness order (a/b/c) & (a) & (b) & (c) & 3H \\
\midrule
MedAgentBench & \textsc{parity} / Terminus-2 / \textsc{fc} & 108/54 & 107/77 & 150/112 & 365/167 \\
Reasoning Gym & Terminus-2 / mini-swe / Goose & 120/86 & 114/90 & 155/119 & 389/143 \\
AppWorld & \textsc{code} / \textsc{fc} / mini-swe & 74/46 & 45/32 & 89/52 & 208/60 \\
BFCL single-turn & \textsc{fc} / \textsc{json} / Terminus-2 & 929/915 & 901/892 & 849/843 & 2,679/952 \\
BFCL multi-turn & \textsc{fc} / \textsc{json} / \textsc{pycall} & 593/374 & 386/279 & 416/282 & 1,395/431 \\
\bottomrule
\end{tabular}
\end{table}

\paragraph{Training-set size controls.}
To test whether the advantage of multi-harness distillation comes simply from using more training trajectories, we compare 3H models with single-harness specialists at matched dataset sizes. For the 108-trajectory comparison, the 3H dataset contains 36 trajectories from each harness and is matched to a 108-trajectory \textsc{parity} specialist. For the 150-trajectory comparison, it contains 50 trajectories from each harness and is matched to a 150-trajectory \textsc{fc} specialist. The 3H subsets are sampled from the full 365-trajectory dataset, either randomly or to maximise coverage of the 209 training tasks. Results are reported in \cref{tab:res_volmatch}.

\begin{table}[htbp]
\centering
\small
\setlength{\tabcolsep}{4pt}
\caption{\textbf{Gains from multi-harness distillation persist at matched training-set size.}
All results use Qwen3-4B. Mean success (\%) with 95\% CIs. On MedAgentBench, 3H models trained on subsets matched to the size of the \textsc{parity}- and \textsc{fc}-only datasets remain above the base model on every harness, avoiding the cross-harness degradation seen with single-harness distillation. At matched size, 3H can approach or exceed specialist performance on the specialist's own harness while substantially improving performance on the others. On Terminus-2, 3H@108 random and 3H@150 coverage-max outperform the full 3H model despite using fewer training trajectories. Coverage-maximised sampling further improves several settings, while full 3H remains strongest overall on \textsc{parity} and \textsc{fc}.}
\label{tab:res_volmatch}
\begin{tabular}{@{}lrrccc@{}}
\toprule
Train set & rows & tasks & \textsc{parity} & Terminus-2 & \textsc{fc} \\
\midrule
Base & --- & --- & 36.1$\pm$9.2 & 3.7$\pm$2.3 & 24.3$\pm$8.4 \\
3H (full) & 365 & 167 & 67.2$\pm$10.2 & 16.9$\pm$5.9 & 55.3$\pm$8.2 \\
\midrule
\textsc{parity}-only & 108 & 54 & 47.3$\pm$11.7 & 4.1$\pm$2.7 & 21.8$\pm$8.6 \\
3H@108, random draw & 108 & 81 & 43.8$\pm$8.7 & 17.8$\pm$5.5 & 36.9$\pm$7.2 \\
3H@108, coverage-max & 108 & 108 & 54.5$\pm$8.8 & 15.9$\pm$5.3 & 34.7$\pm$8.5 \\
\midrule
\textsc{fc}-only & 150 & 112 & 28.7$\pm$11.3 & 0.0 & 45.1$\pm$9.4 \\
3H@150, random draw & 150 & 100 & 63.2$\pm$9.6 & 12.4$\pm$5.0 & 35.0$\pm$8.5 \\
3H@150, coverage-max & 150 & 139 & 52.3$\pm$9.6 & 17.6$\pm$6.3 & 35.1$\pm$8.1 \\
\bottomrule
\end{tabular}
\end{table}

\subsection{Reinforcement Learning}
\label[appendix]{app:rl}

\cref{tab:grpo_hparams} summarises the GRPO configuration, and \cref{tab:rl_budgets} reports the matched rollout counts. Rewards come directly from the benchmark verifier as binary task success, with no shaping for trajectory length, turn count, or termination.

The learning band (\cref{sec:method_rl}) is computed once from the training split and reused across runs. It contains 96 tasks for MAB and 114 for RG. A task is included if, on at least one training harness, the initial checkpoint succeeds on some attempts and fails on others. Tasks that are always solved or always failed on every training harness are excluded. Tasks in learning band are still sampled across all training harnesses, so a particular task--harness pair may produce identical rewards within a GRPO group.

\begin{table}[t]
\centering
\small
\caption{GRPO hyperparameters and training settings shared across RL runs. Single-harness runs use three epochs and multi-harness runs use one so that rollout counts are matched across conditions (\cref{tab:rl_budgets}). Sequence lengths are reported in tokens.}
\label{tab:grpo_hparams}
\begin{tabular}{@{}ll@{}}
\toprule
Setting & Value \\
\midrule
advantage estimator   & GRPO \\
learning rate         & $1\times10^{-6}$ \\
epochs (multi / single) & 1 / 3 \\
train batch (prompts) & 24 \\
samples per prompt    & 6 \\
policy mini-batch     & 24 \\
micro-batch / GPU     & 1 \\
KL penalty            & none \\
loss reduction        & token-mean \\
std.\ normalisation   & off \\
off-policy correction & token-level TIS, clip $2.0$ \\
overlong filtering    & on \\
context / generation  & $24{,}576$ / $8{,}192$ (tokens)\\
trainer strategy      & FSDP, CPU offload \\
placement             & colocated, 2 GPUs \\
inference engine      & vLLM, TP$=2$, util.\ $0.75$ \\
weight sync           & NCCL \\
rollout concurrency   & 24 trials \\
checkpoint interval   & 5 steps \\
\bottomrule
\end{tabular}
\end{table}

\begin{table}[t]
\centering\small
\caption{\textbf{Matched rollout counts for RL.} Single-harness and 3H runs use the same total number of rollouts. Single-harness runs therefore train for three epochs, while 3H runs train for one. Each batch contains 24 task--harness prompts with six samples per prompt. Reasoning Gym step counts vary because of partial batches.}
\label{tab:rl_budgets}
\begin{tabular}{@{}lrrrr@{}}
\toprule
Condition & Tasks $\times$ harnesses & Epochs & Steps & Rollouts \\
\midrule
MedAgentBench, 3H & $96\times3$ & 1 & 12 & 1,728 \\
MedAgentBench, parity-only & $96\times1$ & 3 & 12 & 1,728 \\
Reasoning Gym, 3H & $114\times3$ & 1 & 14 & 2,052 \\
Reasoning Gym, mini-swe-only & $114\times1$ & 3 & 12--15 & 2,052 \\
\bottomrule
\end{tabular}
\end{table}

\section{Rehearsal Details and Controls}
\label[appendix]{app:drill_examples}

For each rehearsal instance, we keep the original environment, verifier, and solution, and replace only the instruction with a mechanical prompt that provides the solution. A demonstrator model then executes the instruction through the unmodified harness, and only trajectories that pass the original verifier are retained.

\paragraph{Rehearsal and controls.}
Rehearsal trajectories are generated by the 3H SFT student checkpoint, except in the explicit Qwen3-32B demonstrator comparison. We construct rehearsal examples programmatically from solved training instances and their metadata, while preserving the original execution environment and verifier. For MAB and RG, rehearsal data replaces an equal number of ordinary training trajectories, so the total training-set size is unchanged. In MAB, we replace either 107 Terminus-2 trajectories or 150 \textsc{fc} trajectories with rehearsal examples, leaving 365 examples in the final pool. In RG, we replace 120 Terminus-2 trajectories with transcription-rehearsal examples, leaving 389 examples. \cref{tab:rehearsal_pool_inventory} reports the resulting pool compositions, and \cref{app:drill_demonstrator} compares the MAB demonstrators. BFCL-MT uses a different construction. Its task-supervision pool contains \textsc{json} and \textsc{pycall} demonstrations while excluding demonstrations in the target \textsc{fc} interface. We then append 268 verified harness-rehearsal traces. These rehearsal examples are constructed so that the target value cannot be copied directly from the instruction: the model must execute the call and recover the value from the returned result. In the main study, we report results for a single target harness, since exhaustively evaluating every possible harness is impractical. To test whether the effect of harness rehearsal is dependent on target harness choice, we also evaluate an XML-rehearsal condition for BFCL-MT (\cref{app:xml_rehearsal}). This setting adds 186 XML rehearsal traces to the full training mixture, allowing us to assess whether the observed rehearsal effect generalises to a different target harness.

\begin{table}[t]
\centering\small
\caption{\textbf{Rehearsal datasets and matched controls.} Counts report the number of trajectories and distinct training tasks. When two datasets appear in the same row, counts follow the same order as their names. Failed-rehearsal control data contain trajectories that did not pass the verifier. Distinct-task counts are unavailable for Reasoning Gym.}
\label{tab:rehearsal_pool_inventory}
\begin{tabular}{@{}llrr@{}}
\toprule
Benchmark & Training data & Trajectories & Tasks \\
\midrule
MedAgentBench & \textsc{parity}{+}\textsc{fc} / \textsc{parity}{+}Terminus-2 & 258 / 215 & 138 / 98 \\
 & rehearsal / failed rehearsal & 107 / 107 & 79 / 70 \\
 & \textsc{parity}{+}\textsc{fc} + rehearsal / 32B rehearsal & 365 / 365 & 168 / 169 \\
 & \textsc{fc} rehearsal / \textsc{parity}{+}Terminus-2 + \textsc{fc} rehearsal & 150 / 365 & 84 / 160 \\
Reasoning Gym & mini-swe{+}Goose / rehearsal & 269 / 120 & 137 / 73 \\
  & mini-swe{+}Goose + rehearsal & 389 & 176 \\
  & failed rehearsal / mini-swe{+}Goose + failed rehearsal & 94 / 363 & --- \\
  & 32B rehearsal / mini-swe{+}Goose + 32B rehearsal & 120 / 389 & --- \\
BFCL multi-turn & task supervision / harness rehearsal & 802 / 268 & 368 / 135 \\
 & task supervision + harness rehearsal & 1,070 & 503 \\
 & task supervision + 38 successful / failed traces & 840 / 840 & 405 / 396 \\
 & 32B rehearsal / task supervision + 32B rehearsal & 235 / 1,037 & 119 / 487 \\
 & 3H + \textsc{xml} rehearsal & 1,581 & 537 \\
\bottomrule
\end{tabular}

\end{table}

\subsection{Example MedAgentBench Harness Rehearsal Instructions}
\label[appendix]{app:drill_mab}

MAB rehearsal uses four templates (task2/3/4/8) instantiated over 96 training examples. We show two representative cases: a read rehearsal (task4), presented in both Terminus-2 and \textsc{fc} form for the same underlying example, and a write rehearsal (task3) that posts a fixed payload. The shared benchmark boilerplate (\texttt{\#\# Environment} and \texttt{\#\# Required Output}), which appears in all MAB instructions, is omitted for brevity.

\bigskip
\noindent\textbf{Read task rehearsal with Terminus-2} (\texttt{task4\_1}):
\vspace{2pt}\hrule\vspace{-4pt}
{\fontsize{7.5pt}{8.8pt}\selectfont\begin{verbatim}

## Procedure
Execute the following steps exactly, in order. Everything you need is
specified below; read command output carefully and transcribe values exactly.

1. Run: `medagentbench_cli.py get "http://localhost:8080/fhir/Observation
   ?patient=S3032536&code=MG&_count=5000&_format=json"`
2. In the returned JSON, consider only entries whose
   `resource.effectiveDateTime` is at or after `2023-11-12T10:15:00+00:00`.
   Among those, find the entry with the LATEST `resource.effectiveDateTime`
   and read V = its `resource.valueQuantity.value` (copy the number exactly
   as it appears).
3. If there is no such entry, run: `medagentbench_cli.py finish '[-1]'`
   Otherwise, run: `medagentbench_cli.py finish '[V]'`
\end{verbatim}
}
\vspace{1pt}\hrule\vspace{6pt}

\noindent\textbf{Read rehearsal with \textsc{fc}} (same instance):
\vspace{2pt}\hrule\vspace{-4pt}
{\fontsize{7.5pt}{8.8pt}\selectfont\begin{verbatim}
## Procedure
Execute the following steps exactly, in order, using the provided tools.
Everything you need is specified below; read tool output carefully and
transcribe values exactly.

1. Call the `get` tool with url `http://localhost:8080/fhir/Observation
   ?patient=S3032536&code=MG&_count=5000&_format=json`.
2. In the returned JSON, consider only entries whose
   `resource.effectiveDateTime` is at or after `2023-11-12T10:15:00+00:00`.
   Among those, find the entry with the LATEST `resource.effectiveDateTime`
   and read V = its `resource.valueQuantity.value` (copy the number exactly
   as it appears).
3. If there is no such entry, call the `finish` tool with answer `[-1]`.
   Otherwise, call the `finish` tool with answer `[V]`.
\end{verbatim}
}
\vspace{1pt}\hrule\vspace{6pt}

\noindent\textbf{Write rehearsal with \textsc{fc}} (\texttt{task3\_1}, where the payload is the ground-truth answer for this training instance and does not transfer across instances):
\vspace{2pt}\hrule\vspace{-4pt}
{\fontsize{7.5pt}{8.8pt}\selectfont\begin{verbatim}
1. Call the `post` tool with url `http://localhost:8080/fhir/Observation`
   and payload exactly:
   `{"resourceType": "Observation", "category": [{"coding": [{"system":
   "http://hl7.org/fhir/observation-category", "code": "vital-signs",
   "display": "Vital Signs"}]}], "code": {"text": "BP"},
   "effectiveDateTime": "2023-11-13T10:15:00+00:00", "status": "final",
   "valueString": "118/77 mmHg",
   "subject": {"reference": "Patient/S2380121"}}`
2. Call the `finish` tool with answer `["DONE"]`.
\end{verbatim}
}
\vspace{1pt}\hrule\vspace{6pt}

\subsection{Example Reasoning Gym Harness Rehearsal Instruction}
\label[appendix]{app:drill_rg}

For RG, harness rehearsal reduces the task to transcription: the model is given the answer string directly, without the original query or schema. Success therefore depends only on executing the required harness operations, including shell quoting, exact file writing, verification, and termination:

\vspace{2pt}\hrule\vspace{-4pt}
{\fontsize{7.5pt}{8.8pt}\selectfont\begin{verbatim}
Transcription rehearsal: your only task is to write an exact string to a file.

Write EXACTLY the following content (between the BEGIN/END markers, not
including the marker lines themselves) to the file "/workspace/answer.txt":

BEGIN
-3.0 - 3.0i
END

Preserve every character, including signs, decimals, spaces and line breaks.
Use careful shell quoting (e.g. cat with a quoted heredoc) so nothing is
interpreted by the shell. Verify the file content matches before finishing.
\end{verbatim}
}
\vspace{1pt}\hrule\vspace{6pt}

\noindent The provided answer (here \texttt{-3.0 - 3.0i}) is the solution to a single training-seed instance from a procedurally generated task family. Evaluation uses a disjoint seed from the same family, so the rehearsal answer does not directly apply to evaluation instances. Because RG rehearsal contains no schema information, the corresponding effect in \cref{tab:res_drills} also argues against schema leakage as the source of the rehearsal gain.

\subsection{Example BFCL Multi-Turn Harness Rehearsal Instruction}
\label[appendix]{app:drill_mt}

BFCL-MT rehearsal uses 138 instances spanning three executable API families (\texttt{GorillaFileSystem}, \texttt{TradingBot}, \texttt{VehicleControlAPI}). The rehearsal instructions are interface-independent: the harness system prompt supplies the action syntax, so the same instance can be executed under \textsc{fc}, \textsc{json}, \textsc{pycall}, or XML without changing the instruction itself. Each instance also includes the benchmark-standard initial configuration and function list. These are omitted below for brevity.

\noindent\textbf{Example rehearsal} (\texttt{drill-gorillafilesystem-0}, involving \texttt{TwitterAPI} and \texttt{GorillaFileSystem}):
\vspace{2pt}\hrule\vspace{-4pt}
{\fontsize{7.5pt}{8.8pt}\selectfont\begin{verbatim}
Turn 1. Call `ls(a=True)` and read its result.

Turn 2. Let V be the FIRST entry listed by that result which does not start
        with a dot. Call `touch(file_name='rehearsal_<V>.txt')', substituting V.
\end{verbatim}
}
\vspace{1pt}\hrule\vspace{6pt}

The target value \(V\) appears only in the environment response to the first call. The model must therefore execute that call, read the result, substitute \(V\), and issue the dependent second call. Rehearsals are discarded if \(V\) appears anywhere in the instruction, preventing solution by copying. The actions themselves are fully specified and only the environment-provided value is unknown.

Verified rehearsal trajectories are collected with the unmodified harness and verifier, retaining only passing traces and at most two per rehearsal. This yields 268 \textsc{fc} traces for \cref{tab:res_drills} and 186 \textsc{xml} traces for \cref{app:xml_rehearsal}.

\begin{table}[t]
\centering
\small
\setlength{\tabcolsep}{4.5pt}
\caption{\textbf{Full MedAgentBench results for Qwen3-4B across all training pools.} Columns are evaluation harnesses. \textsc{xml} is held out from every pool. Rehearsal rows are grouped by the rehearsed target harness, each with cross-harness task supervision from the other training harnesses. Beyond the main-text conditions, the table adds rehearsal built from verifier-failing traces, rehearsal demonstrated by the Qwen3-32B teacher rather than the 3H student (\cref{app:drill_demonstrator}), and the full design repeated with \textsc{fc} as the target. Mean success (\%) with 95\% CIs. The GRPO row is three-harness GRPO from the 3H checkpoint, the mean of two training seeds, as in \cref{tab:res_singlerl}.}
\label{tab:res_qwen_full}
\begin{tabular}{@{}lcccc@{}}
\toprule
 & \multicolumn{4}{c}{Evaluation harness} \\
\cmidrule(l){2-5}
Training pool & \textsc{parity} & Terminus-2 & \textsc{fc} & \textsc{xml} (held-out) \\
\midrule
none (base) & 36.1$\pm$9.2 & 3.7$\pm$2.3 & 24.3$\pm$8.4 & 49.0$\pm$11.1 \\
\midrule
\multicolumn{5}{@{}l}{\textbf{\emph{Single-harness distillation}}} \\
\quad \textsc{parity}   & 47.3$\pm$11.7 & 4.1$\pm$2.7 & 21.8$\pm$8.6 & 43.8$\pm$10.5 \\
\quad Terminus-2        & 26.4$\pm$7.6 & 21.7$\pm$6.4 & 16.1$\pm$5.2 & 32.9$\pm$8.0 \\
\quad \textsc{fc}       & 28.7$\pm$11.3 & 0.0 & 45.1$\pm$9.4 & 58.4$\pm$7.2 \\
\midrule
\multicolumn{5}{@{}l}{\textbf{\emph{Rehearsal target: Terminus-2}} \quad (cross-harness task supervision: \textsc{parity}{+}\textsc{fc})} \\
\quad task supervision only                          & 63.5$\pm$10.0 & 2.0$\pm$1.6 & 47.1$\pm$10.0 & 63.9$\pm$9.0 \\
\quad rehearsal only                           & 40.4$\pm$10.2 & 9.5$\pm$5.7 & 38.5$\pm$10.9 & 38.3$\pm$8.1 \\
\quad rehearsal only, failed traces            & 40.9$\pm$11.0 & 2.1$\pm$1.8 & 32.5$\pm$10.5 & 37.0$\pm$9.0 \\
\quad task supervision{+}rehearsal                   & 52.5$\pm$9.0 & 34.4$\pm$8.2 & 57.0$\pm$9.0 & 49.0$\pm$8.6 \\
\quad task supervision{+}rehearsal, 32B demonstrator & 70.2$\pm$9.5 & 21.7$\pm$7.6 & 50.4$\pm$8.6 & 63.8$\pm$10.8 \\
\midrule
\multicolumn{5}{@{}l}{\textbf{\emph{Rehearsal target: \textsc{fc}}} \quad (cross-harness task supervision: \textsc{parity}{+}Terminus-2)} \\
\quad task supervision only          & 46.7$\pm$9.8 & 18.0$\pm$6.8 & 25.0$\pm$8.4 & 47.1$\pm$10.9 \\
\quad rehearsal only           & 26.6$\pm$11.3 & 0.0 & 39.6$\pm$9.6 & 39.8$\pm$9.4 \\
\quad task supervision{+}rehearsal   & 61.3$\pm$10.2 & 23.0$\pm$8.2 & 36.6$\pm$9.8 & 42.6$\pm$9.6 \\
\midrule
\multicolumn{5}{@{}l}{\textbf{\emph{Three-harness distillation}}} \\
\quad 3H            & 67.2$\pm$10.2 & 16.9$\pm$5.9 & 55.3$\pm$8.2 & 71.5$\pm$10.0 \\
\quad 3H{+}GRPO     & 77.6$\pm$8.4 & 50.8$\pm$9.7 & 85.7$\pm$6.2 & 78.0$\pm$8.1 \\
\bottomrule
\end{tabular}
\end{table}

\subsection{Demonstrator Process Quality}
\label[appendix]{app:drill_demonstrator}

The two task-supervision-plus-rehearsal conditions in the Terminus-2 block of \cref{tab:res_qwen_full} use the same 107 rehearsal instructions, harness, and verifier. They differ only in the demonstrator: the 3H 4B student or the Qwen3-32B teacher. Despite passing the same verifier, their trajectories differ substantially in length. The 4B demonstrator has a median of 17 assistant turns and a maximum of 287, compared with 4 and 6 for the 32B teacher (\cref{tab:app_drillstats}).

\begin{table}[h]
\centering
\caption{\textbf{Execution length of verified rehearsal traces.}}
\label{tab:app_drillstats}
\begin{tabular}{lcc}
\toprule
 & 4B demonstrator & 32B demonstrator \\
\midrule
mean assistant turns & 34.7 & 3.6 \\
median / p90 / max turns & 17 / 69 / 287 & 4 / 5 / 6 \\
mean assistant characters & 30{,}870 & 13{,}622 \\
\bottomrule
\end{tabular}
\end{table}

This difference also changes performance on the unrehearsed \textsc{parity} harness. Relative to task supervision alone ($63.5$), rehearsal generated by the 4B demonstrator lowers \textsc{parity} to $52.5$ ($-11.0$ points), whereas rehearsal generated by the 32B teacher raises it to $70.2$ ($+6.7$ points). Because the rehearsal instructions and task supervision are otherwise unchanged, the comparison indicates that the quality of the demonstrated execution process can affect transfer beyond the rehearsed harness.

\subsection{Rehearsal on a Second Held-Out Harness}
\label[appendix]{app:xml_rehearsal}

Our main BFCL-MT results use \textsc{fc} as the held-out target interface. To test whether the rehearsal effect depends on this choice, we repeat the experiment with XML, which is also excluded from the distillation data (\cref{tab:res_xml_rehearsal}). Three-harness distillation leaves XML performance near the base model ($20.6\!\to\!19.9$), while adding 186 XML rehearsal trajectories raises it to $36.6$. Performance on the three training interfaces changes by at most 1.2 points. The rehearsal effect therefore also appears on a second held-out interface and is concentrated on the rehearsed interface.

\begin{table}[t]
\centering
\small
\setlength{\tabcolsep}{5pt}
\caption{\textbf{BFCL multi-turn rehearsal on a held-out XML interface.} We compare the 3H SFT checkpoint with and without 186 additional \textsc{xml} rehearsal trajectories. XML is excluded from the task supervision data. Mean success (\%) with 95\% CIs on 200 test tasks. Bold marks the best result on the target harness. The final row reports the change from the 3H control.}
\label{tab:res_xml_rehearsal}
\begin{tabular}{@{}lcccc@{}}
\toprule
Training pool & \textsc{xml} (target) & \textsc{fc} & \textsc{json} & \textsc{pycall} \\
\midrule
None (Base)                 & 20.6$\pm$3.8 & 44.6$\pm$5.9 & 33.5$\pm$5.4 & 28.2$\pm$5.1 \\
3H (control)                & 19.9$\pm$4.2 & 59.0$\pm$5.5 & 49.5$\pm$5.4 & 49.1$\pm$5.6 \\
3H + \textsc{xml} rehearsal & \textbf{36.6$\pm$4.8} & 60.2$\pm$5.8 & 49.9$\pm$5.2 & 48.9$\pm$5.9 \\
\midrule
$\Delta$ vs control         & $+16.7$ & $+1.2$ & $+0.4$ & $-0.2$ \\
\bottomrule
\end{tabular}
\end{table}

\section{Extended Results}
\label[appendix]{sec:appendix_results}

\subsection{Tasks Solved Across Harnesses and Models}
\label[appendix]{sec:appendix_reach}

\cref{tab:res_reach} complements \cref{sec:res_exposure} to every model and benchmark we evaluated.

\begin{table}[t]
\centering
\small
\setlength{\tabcolsep}{4.5pt}
\caption{\textbf{Different harnesses expose complementary capabilities.} Harnesses differ not only in overall performance but also in which tasks they enable the policy to solve. \emph{Solved by any} counts tasks solved through at least one harness, while \emph{Strongest harness} reports the single harness solving the most tasks. \emph{Others add} counts tasks missed by the strongest harness but solved through another harness. The percentage is the share of all solved tasks contributed uniquely by these other harnesses. $^\dagger$ denotes results over four harnesses, including the held-out harness.}
\label{tab:res_reach}
\begin{tabular}{@{}llrrrlr@{}}
\toprule
Benchmark & Model & Attempts & Tasks & Solved by any & Strongest harness & Others add \\
\midrule
MedAgentBench & Qwen3-4B & pass@4 & 61 & 46 & \textsc{parity} (38) & $+8$ (17\%) \\
 & Qwen3-32B & pass@4 & 61 & 53 & \textsc{parity} (50) & $+3$ (6\%) \\
 & Gemma-3-4B$^\dagger$ & pass@4 & 61 & 23 & \textsc{parity} (16) & $+7$ (30\%) \\
\midrule
Reasoning Gym (easy) & Qwen3-4B & pass@4 & 96 & 75 & mini-swe (71) & $+4$ (5\%) \\
 & Qwen3-32B & pass@4 & 96 & 82 & mini-swe (77) & $+5$ (6\%) \\
 & Gemma-3-4B$^\dagger$ & pass@4 & 96 & 31 & Terminus-2 (23) & $+8$ (26\%) \\
\midrule
Reasoning Gym (hard) & Qwen3-4B & pass@4 & 96 & 61 & mini-swe (58) & $+3$ (5\%) \\
 & Qwen3-32B & pass@4 & 96 & 79 & mini-swe (74) & $+5$ (6\%) \\
\midrule
BFCL multi-turn & Qwen3-4B & pass@4 & 200 & 141 & \textsc{fc} (125) & $+16$ (11\%) \\
 & Qwen3-32B & greedy & 200 & 131 & \textsc{fc} (116) & $+15$ (11\%) \\
\midrule
AppWorld & Qwen3-4B & pass@4 & 168 & 16 & \textsc{fc} (14) & $+2$ (12\%) \\
 & Qwen3-32B & greedy & 168 & 89 & mini-swe (49) & $+40$ (45\%) \\
\bottomrule
\end{tabular}
\end{table}

\subsection{Distillation and Held-Out Transfer}

\cref{tab:res_ladders_app} completes \cref{tab:res_ladders_main}, and \cref{tab:res_heldout} gives the full held-out scores behind \cref{fig:res_heldout}.

\begin{table}[t]
\centering
\small
\setlength{\tabcolsep}{2pt}
\caption{\textbf{Additional distillation results for BFCL single-turn and Reasoning Gym Easy} Mean success (\%) with 95\% CIs. BFCL single-turn uses one greedy attempt and has no interval. Arrows show change from the base model, and bold indicates the best result in each column. The Reasoning Gym Easy block over four training runs is in \cref{tab:reb_D}.}
\label{tab:res_ladders_app}
\begin{tabular}{@{}lccc @{\hspace{5pt}} lccc@{}}
\toprule
\multicolumn{4}{c}{\textbf{BFCL single-turn}} & \multicolumn{4}{c}{\textbf{Reasoning Gym Easy}} \\
\cmidrule(r{6pt}){1-4}\cmidrule(l){5-8}
Training & \textsc{fc} & \textsc{json} & Terminus-2 & Training & Terminus-2 & mini-swe & Goose \\
\midrule
Base & \textbf{87.0} & 86.2 & 22.5 & Base & 18.0$\pm$5.5 & 59.4$\pm$7.9 & 25.7$\pm$6.7 \\
\textsc{fc}-only & 85.1\dn & 21.4\dn & 0.0\dn & Terminus-2-only & 54.7$\pm$7.8\up & 55.7$\pm$8.4\dn & 40.2$\pm$7.3\up \\
\textsc{json}-only & 85.1\dn & \textbf{87.9}\up & 1.5\dn & mini-swe-only & 18.4$\pm$4.4\up & 44.2$\pm$7.7\dn & 39.8$\pm$7.4\up \\
Terminus-2-only & 21.4\dn & 83.9\dn & 42.5\up & Goose-only & 37.3$\pm$6.7\up & 51.5$\pm$7.7\dn & \textbf{64.6$\pm$8.1}\up \\
\rowcolor{gray!12}
3H & 86.8\dn & 86.2\nc & \textbf{43.2}\up & 3H & \textbf{55.3$\pm$7.9}\up & \textbf{63.0$\pm$7.6}\up & 60.8$\pm$8.2\up \\
\bottomrule
\end{tabular}
\end{table}

\begin{table}[!ht]
\centering
\small
\setlength{\tabcolsep}{4pt}
\caption{\textbf{Transfer to a held-out harness.} Results corresponding to the held-out-transfer experiments shown in \cref{fig:res_heldout}. We evaluate the base model, three single-harness specialists, and the 3H SFT checkpoint on a fourth harness excluded from training. Mean success (\%) with 95\% CIs. BFCL single-turn uses one greedy attempt and has no interval. Arrows indicate change from the base model. Bold indicates the best-performing trained model in each row. Held-out results over multiple training runs are in \cref{tab:reb_D}.}
\label{tab:res_heldout}
\begin{tabular}{@{}llccccc@{}}
\toprule
& held-out & & \multicolumn{3}{c}{single-harness} & \\
\cmidrule(lr){4-6}
Benchmark & harness & base & (a) & (b) & (c) & 3H \\
\midrule
MedAgentBench              & \textsc{xml}   & 49.0$\pm$11.1 & 43.8$\pm$10.5\dn & 32.9$\pm$8.0\dn & 58.4$\pm$7.2\up & \textbf{71.5$\pm$10.2}\up \\
AppWorld         & \textsc{xml}   & 4.5$\pm$3.8 & 9.5$\pm$4.3\up & 3.3$\pm$2.2\dn & 14.1$\pm$5.0\up & \textbf{21.3$\pm$7.6}\up \\
BFCL multi-turn  & \textsc{xml}   & 20.6$\pm$3.8 & 2.5$\pm$1.1\dn & \textbf{20.0$\pm$4.1}\dn & 18.8$\pm$3.9\dn & 19.9$\pm$4.2\dn \\
BFCL single-turn & \textsc{xml}   & 86.2 & 21.7\dn & \textbf{88.5}\up & 21.4\dn & 87.0\up \\
Reasoning Gym Easy          & Aider & 20.8$\pm$6.8 & 2.4$\pm$2.3\dn & 0.0\dn & 0.0\dn & \textbf{19.0$\pm$6.0}\dn \\
Reasoning Gym Hard          & Aider & 17.6$\pm$6.8 & 0.5$\pm$0.7\dn & 0.0\dn & 0.2$\pm$0.2\dn & \textbf{9.8$\pm$4.8}\dn \\
\bottomrule
\end{tabular}

\vspace{3pt}

\par\smallskip
{\footnotesize\raggedright Specialist harness order (a/b/c): MAB, \textsc{parity}/Terminus-2/\textsc{fc}. AppWorld, \textsc{code}/\textsc{fc}/mini-swe. BFCL-MT, \textsc{fc}/\textsc{json}/\textsc{pycall}. BFCL-ST, \textsc{fc}/\textsc{json}/Terminus-2. RG, Terminus-2/mini-swe/Goose.\par}
\end{table}

\subsection{MedAgentBench: Full Results}

\cref{tab:res_qwen_full} reports every MAB condition for Qwen3-4B. The Gemma-3-4B-it counterpart is in \cref{app:gemma}.

\subsection{Replication with a Second Model Family}
\label[appendix]{app:gemma}

We also report results with Gemma-3-4B-it as the student. \cref{tab:res_gemma_full} reports MAB distillation, rehearsal, and GRPO results, and \cref{tab:res_gemma_rg} reports RG distillation results. The main patterns seen with Qwen3-4B replicate with Gemma-3-4B-it: single-harness distillation often degrades other harnesses, 3H training performs strongly across training harnesses, rehearsal improves the target harness, and GRPO again benefits the weakest training harness. However, on RG, Gemma does not preserve performance on the held-out Aider harness after 3H distillation. This suggests that held-out transfer can depend on the model family. Under GRPO (two training seeds each), Gemma also mirrors the Qwen3-4B RL results: 3H GRPO improves every MAB harness over 3H SFT, including the held-out \textsc{xml} ($44.3\to53.0$), and \textsc{parity}-only GRPO raises \textsc{parity}, Terminus-2 and \textsc{xml} without a significant drop on \textsc{fc}.

\begin{table}[t]
\centering
\small
\setlength{\tabcolsep}{4.5pt}
\caption{\textbf{Reasoning Gym Easy results with Gemma-3-4B-it.} We report single- and multi-harness distillation results on the three training harnesses and the held-out Aider harness. The 3H setting remains strongest across the training harnesses, consistent with transfer of task competence across interfaces, while held-out transfer differs from the Qwen results and may be more sensitive to model choice. Mean success (\%) with 95\% CIs. Bold marks the best result in each column.}
\label{tab:res_gemma_rg}
\begin{tabular}{@{}lcccc@{}}
\toprule
Pool & Terminus-2 & mini-swe & Goose & Aider (held out) \\
\midrule
none (base)         & 11.7$\pm$4.4 & 0.0 & 0.0 & \textbf{16.4$\pm$6.5} \\
\midrule
Terminus-2-only    & 26.1$\pm$7.0 & 0.0 & 0.0 & 8.3$\pm$5.1 \\
mini-swe-only   & 0.8$\pm$0.8 & 30.1$\pm$7.5 & 0.0 & 1.0$\pm$1.6 \\
Goose-only & 3.2$\pm$1.8 & 1.6$\pm$1.8 & 0.8$\pm$0.9 & 1.0$\pm$1.6 \\
\midrule
3H & \textbf{27.6$\pm$6.5} & \textbf{32.0$\pm$7.5} & \textbf{19.9$\pm$5.8} & 0.0 \\
\bottomrule
\end{tabular}
\end{table}

\begin{table}[t]
\centering
\small
\setlength{\tabcolsep}{4.5pt}
\caption{\textbf{MedAgentBench results with Gemma-3-4B-it.} Results include all main experimental settings including distillation, harness rehearsal, and GRPO conditions. \textsc{xml} is the held-out harness, and row blocks mirror \cref{tab:res_qwen_full}. The Terminus-2 rehearsal block uses \textsc{parity} alone for cross-harness task supervision. Mean success (\%) with 95\% CIs. For the SFT rows, the \textsc{parity} harness parses the action after the model's reasoning block, as the reasoning parser does for Qwen3. GRPO rows are three-harness and \textsc{parity}-only GRPO from the 3H checkpoint (\cref{sec:res_rl}), each the mean of two training seeds. Bold marks the best result in each column.}
\label{tab:res_gemma_full}
\begin{tabular}{@{}lcccc@{}}
\toprule
 & \multicolumn{4}{c}{Evaluation harness} \\
\cmidrule(l){2-5}
Training pool & \textsc{parity} & Terminus-2 & \textsc{fc} & \textsc{xml} (held-out) \\
\midrule
none (base) & 20.2$\pm$9.4 & 3.7$\pm$2.7 & 26.2$\pm$11.5 & 15.2$\pm$8.0 \\

\midrule
\multicolumn{5}{@{}l}{\textbf{\emph{Single-harness distillation}}} \\
\quad \textsc{parity}   & 31.3$\pm$9.0 & 0.0 & 26.5$\pm$11.5 & 23.0$\pm$8.8 \\
\quad Terminus-2        & 26.2$\pm$11.1 & 21.1$\pm$7.6 & 26.7$\pm$11.5 & 14.2$\pm$6.0 \\
\quad \textsc{fc}       & 26.2$\pm$11.5 & 0.0 & 58.8$\pm$9.2 & 35.0$\pm$8.2 \\
\midrule
\multicolumn{5}{@{}l}{\textbf{\emph{Rehearsal target: Terminus-2}} \quad (cross-harness task supervision: \textsc{parity}, row above)} \\
\quad rehearsal only                 & 27.4$\pm$11.2 & 7.8$\pm$4.1 & 24.8$\pm$11.5 & 20.2$\pm$8.6 \\
\quad rehearsal only, failed traces  & 20.0$\pm$9.2 & 2.5$\pm$2.3 & 25.7$\pm$11.5 & 16.2$\pm$7.2 \\
\quad task supervision{+}rehearsal         & 38.8$\pm$8.7 & 14.1$\pm$6.1 & 25.2$\pm$11.5 & 23.4$\pm$8.2 \\
\midrule
\multicolumn{5}{@{}l}{\textbf{\emph{Rehearsal target: \textsc{fc}}} \quad (cross-harness task supervision: \textsc{parity}{+}Terminus-2)} \\
\quad task supervision only          & 32.5$\pm$9.7 & 20.5$\pm$7.2 & 25.9$\pm$11.5 & 22.6$\pm$7.5 \\
\quad rehearsal only           & 26.3$\pm$11.5 & 0.0 & 17.6$\pm$6.3 & 24.8$\pm$9.1 \\
\quad task supervision{+}rehearsal   & 38.6$\pm$10.5 & 28.8$\pm$8.3 & 42.6$\pm$8.1 & 40.7$\pm$9.4 \\
\midrule
\multicolumn{5}{@{}l}{\textbf{\emph{Three-harness distillation}}} \\
\quad 3H            & 55.4$\pm$11.3 & 20.2$\pm$6.6 & 73.5$\pm$8.0 & 44.3$\pm$9.4 \\
\quad 3H{+}GRPO     & 66.5$\pm$9.7 & \textbf{40.2$\pm$8.9} & \textbf{77.2$\pm$7.8} & \textbf{53.0$\pm$8.4} \\
\quad 3H{+}\textsc{parity}-only GRPO & \textbf{69.1$\pm$9.5} & 37.0$\pm$8.3 & 71.8$\pm$8.8 & 47.8$\pm$8.4 \\
\bottomrule
\end{tabular}
\end{table}

\section{Additional Controls and Robustness Analyses}
\label[appendix]{app:rebuttal_controls}

This appendix reports every main training condition over four independent training runs, matched-size and matched-budget controls, controls for task leakage through rehearsal, and a failure-type analysis of what rehearsal changes. All use Qwen3-4B and the evaluation protocol of \cref{sec:appendix_eval}.

\subsection{Robustness Across Training Runs}
\label[appendix]{app:seeds}
\begin{table}[t]
\centering
\footnotesize
\setlength{\tabcolsep}{5pt}
\renewcommand{\arraystretch}{1.02}

\caption{\textbf{The rehearsal $2\times2$ on the target harness over four training runs.}
Columns report the base model, task supervision with cross-harness distillation (\emph{Cross-task}), target-harness rehearsal (\emph{Rehearsal}), their combination, direct task supervision through the target-harness specialist distillation (\emph{Target-task}), and multi-harness distillation through all three training harnesses (\emph{3H}). Cross-harness task supervision transfers poorly on its own, while rehearsal alone provides limited task learning. Together, they substantially improve target-harness performance. Each trained cell is the equal-weight mean of four independent training runs (s0--s3) with a 95\% task-bootstrap CI of that mean. Per-run values are in \cref{tab:reb_N}, and \cref{tab:res_drills} in the main text reports run s0 on every evaluation harness. Bold marks the best of the three rehearsal-design columns in each row.}
\label{tab:res_drills_pooled}
\begin{tabular}{@{}lcccccc@{}}
\toprule
Harness
& Base
& Cross-task
& Rehearsal
& \makecell{Cross-task\\+ rehearsal}
& Target-task
& 3H \\
\midrule

\multicolumn{7}{@{}l}{\textbf{MedAgentBench} \quad target: Terminus-2} \\
Terminus-2
& 5.0$\pm$2.8
& 6.2$\pm$2.6
& 12.3$\pm$5.3
& \textbf{40.5$\pm$7.7}
& 25.3$\pm$7.2
& 25.1$\pm$6.6 \\

\midrule
\multicolumn{7}{@{}l}{\textbf{Reasoning Gym Easy} \quad target: Terminus-2} \\
Terminus-2
& 18.8$\pm$5.3
& 20.2$\pm$3.6
& 19.7$\pm$4.7
& \textbf{39.1$\pm$6.5}
& 52.9$\pm$7.9
& 56.6$\pm$7.5 \\

\midrule
\multicolumn{7}{@{}l}{\textbf{Reasoning Gym Hard} \quad target: Terminus-2} \\
Terminus-2
& 16.4$\pm$5.5
& 11.9$\pm$3.3
& 14.3$\pm$4.1
& \textbf{28.8$\pm$5.5}
& 40.2$\pm$7.6
& 39.5$\pm$7.1 \\

\midrule
\multicolumn{7}{@{}l}{\textbf{BFCL multi-turn} \quad target: \textsc{fc}} \\
\textsc{fc}
& 44.6$\pm$5.9
& 12.8$\pm$2.9
& 31.1$\pm$4.6
& \textbf{56.8$\pm$5.0}
& 53.8$\pm$4.9
& 60.2$\pm$5.1 \\

\bottomrule
\end{tabular}
\end{table}

We train every target-harness arm of the rehearsal $2\times2$ and of the distillation ladders with four independent seeds, with the same data, recipe and evaluation protocol. \cref{tab:res_drills_pooled} gives the rehearsal $2\times2$ on the target harness averaged over the four runs and \cref{tab:reb_N} each run individually, and \cref{tab:reb_D} the distillation ladders over four runs behind \cref{tab:res_ladders_main}. The combined arm is the best arm in every run on every benchmark, by at least 10 points, and 3H avoids the held-out collapses of the specialists in every run. Run-to-run variance is concentrated in BFCL-MT task supervision (8.5 / 8.6 / 26.0 / 8.2 across the four runs, and 23.6 in the run that serves as the reference for the controls in \cref{tab:reb_A,tab:reb_E}). This instability is itself an interface effect: in the low runs about 40\% of assistant turns write the tool call as text in the \textsc{json} or \textsc{pycall} format seen in training instead of making a native \textsc{fc} call, and 26--30\% of trials never make a native call, against about 20\% of turns and 6--12\% of trials in the high runs. With target-harness rehearsal this variation largely disappears: the combined arm stays at 54.9--58.8 across all four runs and text calls fall to about 1\% of turns.

\begin{table}[htbp]
\centering\footnotesize\setlength{\tabcolsep}{4pt}

\caption{\textbf{Rehearsal 2$\times$2 per training run: target-harness score (\%)} Mean $\pm$ 95\% task-bootstrap CI, 4 attempts per task. For the benchmarks in \cref{tab:res_drills}, run s0 is the run shown there. \cref{tab:res_drills_pooled} reports the equal-weight mean of the four runs.}
\label{tab:reb_N}
\begin{tabular}{@{}lcccc@{}}
\toprule
Benchmark & Run & Task only & Rehearsal only & Task + rehearsal \\
\midrule
BFCL multi-turn (\textsc{fc}) & s0 & 8.5 $\pm$ 2.8 & 32.5 $\pm$ 5.1 & \textbf{58.8 $\pm$ 5.7} \\
 & s1 & 8.6 $\pm$ 2.9 & 32.2 $\pm$ 4.8 & \textbf{54.9 $\pm$ 5.2} \\
 & s2 & 26.0 $\pm$ 4.9 & 28.9 $\pm$ 5.0 & \textbf{56.1 $\pm$ 5.4} \\
 & s3 & 8.2 $\pm$ 3.1 & 30.8 $\pm$ 5.1 & \textbf{57.5 $\pm$ 5.4} \\
Reasoning Gym easy (Terminus-2) & s0 & 19.8 $\pm$ 5.9 & 20.0 $\pm$ 5.7 & \textbf{37.0 $\pm$ 7.1} \\
 & s1 & 24.9 $\pm$ 5.8 & 24.8 $\pm$ 5.9 & \textbf{38.9 $\pm$ 7.5} \\
 & s2 & 20.0 $\pm$ 5.2 & 16.5 $\pm$ 5.0 & \textbf{40.2 $\pm$ 7.3} \\
 & s3 & 16.1 $\pm$ 4.6 & 17.7 $\pm$ 4.8 & \textbf{40.1 $\pm$ 7.8} \\
Reasoning Gym hard (Terminus-2) & s0 & 15.1 $\pm$ 4.9 & 14.7 $\pm$ 5.0 & \textbf{28.5 $\pm$ 6.1} \\
 & s1 & 6.6 $\pm$ 2.9 & 16.4 $\pm$ 5.2 & \textbf{26.7 $\pm$ 6.2} \\
 & s2 & 12.6 $\pm$ 4.0 & 12.8 $\pm$ 4.3 & \textbf{28.5 $\pm$ 6.4} \\
 & s3 & 13.3 $\pm$ 4.3 & 13.3 $\pm$ 4.8 & \textbf{31.6 $\pm$ 7.1} \\
MedAgentBench (Terminus-2) & s0 & 2.0 $\pm$ 1.6 & 9.5 $\pm$ 5.7 & \textbf{34.4 $\pm$ 8.2} \\
 & s1 & 7.4 $\pm$ 3.9 & 12.9 $\pm$ 7.2 & \textbf{43.9 $\pm$ 9.4} \\
 & s2 & 10.7 $\pm$ 4.5 & 16.1 $\pm$ 5.9 & \textbf{37.9 $\pm$ 9.4} \\
 & s3 & 4.6 $\pm$ 3.3 & 10.7 $\pm$ 4.4 & \textbf{45.9 $\pm$ 8.6} \\
\bottomrule
\end{tabular}\par

\end{table}
\begin{table}[htbp]
\centering\footnotesize\setlength{\tabcolsep}{4pt}

\caption{\textbf{Distillation results over four training runs: score (\%) per evaluation harness (held-out harness last)}}
\label{tab:reb_D}
\par\medskip\noindent\emph{MedAgentBench}\par\smallskip
\begin{tabular}{@{}lcccc@{}}
\toprule
 & \textsc{parity} & Terminus-2 & \textsc{fc} & \textsc{xml} (held out) \\
\midrule
parity-only & 45.8 $\pm$ 10.3 & 7.6 $\pm$ 3.4 & 18.6 $\pm$ 7.8 & 46.3 $\pm$ 9.3 \\
Terminus-2-only & 28.8 $\pm$ 7.7 & \textbf{25.3 $\pm$ 7.2} & 21.7 $\pm$ 6.8 & 28.5 $\pm$ 7.2 \\
fc-only & 28.5 $\pm$ 11.2 & 0.0 $\pm$ 0.0 & 48.3 $\pm$ 7.9 & 55.5 $\pm$ 8.0 \\
3H & \textbf{68.0 $\pm$ 10.3} & 25.1 $\pm$ 6.6 & \textbf{52.4 $\pm$ 7.2} & \textbf{68.7 $\pm$ 10.0} \\
\bottomrule
\end{tabular}\par
\par\medskip\noindent\emph{Reasoning Gym, easy}\par\smallskip
\begin{tabular}{@{}lcccc@{}}
\toprule
 & Terminus-2 & mini-swe & Goose & Aider (held out) \\
\midrule
Terminus-2-only & 52.9 $\pm$ 7.9 & 57.1 $\pm$ 7.6 & 26.8 $\pm$ 5.3 & 2.4 $\pm$ 2.0 \\
mini-swe-only & 20.7 $\pm$ 4.0 & 48.8 $\pm$ 7.9 & 49.9 $\pm$ 7.9 & 0.0 $\pm$ 0.0 \\
Goose-only & 40.1 $\pm$ 6.8 & 53.0 $\pm$ 7.8 & 61.3 $\pm$ 8.4 & 0.8 $\pm$ 0.9 \\
3H & \textbf{56.6 $\pm$ 7.5} & \textbf{61.4 $\pm$ 7.3} & \textbf{63.8 $\pm$ 8.3} & \textbf{13.1 $\pm$ 4.3} \\
\bottomrule
\end{tabular}\par
\par\medskip\noindent\emph{Reasoning Gym, hard}\par\smallskip
\begin{tabular}{@{}lcccc@{}}
\toprule
 & Terminus-2 & mini-swe & Goose & Aider (held out) \\
\midrule
Terminus-2-only & \textbf{40.2 $\pm$ 7.6} & 41.5 $\pm$ 7.8 & 23.8 $\pm$ 5.6 & 1.3 $\pm$ 1.0 \\
mini-swe-only & 15.6 $\pm$ 4.2 & 35.0 $\pm$ 7.6 & 31.7 $\pm$ 7.2 & 0.0 $\pm$ 0.0 \\
Goose-only & 28.0 $\pm$ 6.1 & 41.1 $\pm$ 7.2 & 45.4 $\pm$ 8.2 & 0.3 $\pm$ 0.4 \\
3H & 39.5 $\pm$ 7.1 & \textbf{48.1 $\pm$ 8.0} & \textbf{45.8 $\pm$ 8.0} & \textbf{7.0 $\pm$ 3.1} \\
\bottomrule
\end{tabular}\par
\par\medskip\noindent\emph{BFCL multi-turn}\par\smallskip
\begin{tabular}{@{}lcccc@{}}
\toprule
 & \textsc{fc} & \textsc{json} & \textsc{pycall} & \textsc{xml} (held out) \\
\midrule
fc-only & 53.8 $\pm$ 4.9 & 0.1 $\pm$ 0.2 & 14.2 $\pm$ 3.0 & 1.3 $\pm$ 0.4 \\
json-only & 23.4 $\pm$ 3.9 & 47.7 $\pm$ 4.8 & 30.7 $\pm$ 3.9 & 15.8 $\pm$ 3.3 \\
pycall-only & 31.5 $\pm$ 4.3 & 33.8 $\pm$ 4.6 & 47.7 $\pm$ 4.9 & 17.8 $\pm$ 3.4 \\
3H & \textbf{60.2 $\pm$ 5.1} & \textbf{50.2 $\pm$ 5.2} & \textbf{49.1 $\pm$ 4.8} & \textbf{20.5 $\pm$ 3.7} \\
\bottomrule
\end{tabular}\par
\par\medskip\noindent\emph{AppWorld}\par\smallskip
\begin{tabular}{@{}lcccc@{}}
\toprule
 & \textsc{code} & \textsc{fc} & mini-swe & \textsc{xml} (held out) \\
\midrule
code-only & 12.8 $\pm$ 3.1 & 9.2 $\pm$ 2.7 & 7.8 $\pm$ 2.2 & 10.9 $\pm$ 2.7 \\
fc-only & 2.4 $\pm$ 0.8 & 1.7 $\pm$ 0.7 & 0.9 $\pm$ 0.5 & 3.5 $\pm$ 1.2 \\
mini-swe-only & 14.2 $\pm$ 3.1 & 7.1 $\pm$ 2.0 & 12.6 $\pm$ 3.1 & 15.6 $\pm$ 3.3 \\
3H & \textbf{19.5 $\pm$ 3.8} & \textbf{16.2 $\pm$ 3.6} & \textbf{20.9 $\pm$ 4.1} & \textbf{20.7 $\pm$ 4.1} \\
\bottomrule
\end{tabular}\par

\end{table}

\subsection{Size- and Budget-Matched $2\times2$}
\cref{tab:reb_E} matches the combined pool to the task-supervision arm's size in both directions: reducing the combined pool to the task-supervision arm's row count (keeping all rehearsal rows, or keeping the original mixture) and repeating task supervision up to the combined size. Reduced to the task arm's size, the combined pool scores within about three points of the full combined pool on every benchmark under both schemes, and repeating task supervision to the combined size gives no gain on any benchmark.

\begin{table}[htbp]
\centering\footnotesize\setlength{\tabcolsep}{4pt}

\caption{\textbf{Size- and budget-matched 2$\times$2: target-harness score (\%)}}
\label{tab:reb_E}
\begin{tabular}{@{}lcccc@{}}
\toprule
 & \makecell{BFCL multi-turn\\(\textsc{fc})} & \makecell{MedAgentBench\\(Terminus-2)} & \makecell{Reasoning Gym\\easy (Terminus-2)} & \makecell{Reasoning Gym\\hard (Terminus-2)} \\
\midrule
Task supervision only & 23.6 $\pm$ 4.7 & 8.6 $\pm$ 4.3 & 18.7 $\pm$ 3.7 & 13.5 $\pm$ 3.3 \\
\addlinespace[3pt]
Rehearsal only & 33.6 $\pm$ 5.1 & 12.9 $\pm$ 7.2 & 18.5 $\pm$ 4.5 & 14.6 $\pm$ 4.1 \\
\addlinespace[3pt]
Task supervision + rehearsal & 57.5 $\pm$ 5.6 & 35.2 $\pm$ 8.4 & 41.4 $\pm$ 6.8 & 30.2 $\pm$ 6.0 \\
\addlinespace[3pt]
\makecell[l]{Combined, reduced to task size,\\all rehearsal rows kept} & 56.2 $\pm$ 5.6 & 33.6 $\pm$ 8.4 & 40.2 $\pm$ 6.4 & 27.9 $\pm$ 6.2 \\
\addlinespace[3pt]
\makecell[l]{Combined, reduced to task size,\\same mixture} & 57.1 $\pm$ 5.2 & 32.0 $\pm$ 7.6 & 42.2 $\pm$ 6.7 & 30.7 $\pm$ 6.2 \\
\addlinespace[3pt]
\makecell[l]{Task only, repeated to\\combined size} & 7.8 $\pm$ 2.8 & 8.6 $\pm$ 4.1 & 21.2 $\pm$ 3.7 & 16.4 $\pm$ 3.1 \\
\bottomrule
\end{tabular}\par

\end{table}

\subsection{Rehearsal Through a Non-Target Harness}
\cref{tab:reb_B} pairs task supervision with the same amount of rehearsal through a non-target training harness (\textsc{json} on BFCL-MT, \textsc{fc} on MAB). Non-target rehearsal falls below task supervision alone on both benchmarks, so the benefit comes from practice on the target interface rather than from rehearsal traces in general.

\begin{table}[htbp]
\centering\footnotesize\setlength{\tabcolsep}{4pt}

\caption{\textbf{Rehearsal through a non-target harness: target-harness success (\%)}}
\label{tab:reb_B}
\begin{tabular}{@{}lcc@{}}
\toprule
 & \makecell{BFCL multi-turn,\\target \textsc{fc}} & \makecell{MedAgentBench,\\target Terminus-2} \\
\midrule
Task supervision only & 23.6 $\pm$ 4.7 & 8.6 $\pm$ 4.3 \\
Task supervision + target-harness rehearsal & \textbf{57.5 $\pm$ 5.6} & \textbf{35.2 $\pm$ 8.4} \\
\makecell[l]{Task supervision + the same amount of\\rehearsal through a non-target training harness} & 13.5 $\pm$ 4.0 & 2.5 $\pm$ 2.0 \\
\bottomrule
\end{tabular}\par

\end{table}

\subsection{Rehearsal with Verifier-Failing Traces}
Keeping only verifier-passing rehearsal traces controls for the quality of harness use but also selects for correct task completion. \cref{tab:reb_A} therefore repeats task + rehearsal with the traces that failed the verifier (38 for BFCL-MT, 94 for RG, 107 for MAB, with the passing pool subsampled to the same size on MAB and BFCL-MT). If rehearsal helps mainly by teaching the interface, failed traces should help when they use the harness well and not when the interaction itself is poor. \cref{tab:reb_H} shows this is what separates the benchmarks: RG failures are mostly valid interactions that differ from successes mainly in the final answer and keep about 80\% of the gain. On BFCL-MT fewer than half are fully valid and keep less than half of the gain. MAB failures are long, broken sessions that often time out and keep almost none. Success filtering therefore mainly removes poor interface demonstrations.

\begin{table}[htbp]
\centering\footnotesize\setlength{\tabcolsep}{4pt}

\caption{\textbf{Failed-rehearsal control: target-harness score (\%)}}
\label{tab:reb_A}
\begin{tabular}{@{}lcccc@{}}
\toprule
 & \makecell{BFCL multi-turn\\(\textsc{fc})} & \makecell{Reasoning Gym\\easy (Terminus-2)} & \makecell{Reasoning Gym\\hard (Terminus-2)} & \makecell{MedAgentBench\\(Terminus-2)} \\
\midrule
Task supervision only & 23.6 $\pm$ 4.7 & 18.7 $\pm$ 3.7 & 13.5 $\pm$ 3.3 & 8.6 $\pm$ 4.3 \\
\addlinespace[3pt]
\makecell[l]{Task + rehearsal\\(verifier-passing traces)} & \textbf{56.8 $\pm$ 5.6} & \textbf{41.4 $\pm$ 6.8} & \textbf{30.2 $\pm$ 6.0} & \textbf{35.2 $\pm$ 8.4} \\
\addlinespace[3pt]
\makecell[l]{Task + rehearsal\\(verifier-failing traces)} & 37.4 $\pm$ 4.9 & 36.8 $\pm$ 6.6 & 26.5 $\pm$ 5.9 & 10.8 $\pm$ 4.6 \\
\addlinespace[3pt]
Share of the gain kept (derived) & 42\% & 80\% & 78\% & 8\% \\
\bottomrule
\end{tabular}\par

\end{table}
\begin{table}[htbp]
\centering\footnotesize\setlength{\tabcolsep}{4pt}

\caption{\textbf{Rehearsal traces, failed / successful}}
\label{tab:reb_H}
\begin{tabular}{@{}lccc@{}}
\toprule
 & \makecell{BFCL multi-turn\\(\textsc{fc})} & \makecell{Reasoning Gym\\(Terminus-2)} & \makecell{MedAgentBench\\(Terminus-2)} \\
\midrule
Traces with every turn valid & 47\% / 100\% & 68\% / 87\% & 6\% / 9\% \\
Traces with no valid turn & 29\% / 0\% & 0\% / 0\% & 0\% / 0\% \\
Median assistant turns & 4 / 4 & 2 / 2 & 56 / 17 \\
\bottomrule
\end{tabular}\par

\end{table}

\subsection{When Task and Interface Supervision Are Complementary}
\cref{tab:reb_G} compares rehearsal alone with two models trained with both task and target-interface supervision, the target-harness specialist and 3H SFT, on run s0. Where the interaction in \cref{tab:res_drills} appears, rehearsal alone reaches only about 35--60\% of these models. On MAB with \textsc{fc} as the target, rehearsal alone already reaches 72--88\% of them, and adding cross-harness task supervision gives no further gain (\cref{tab:res_qwen_full}). The same holds on RG when rehearsal traces come from the 32B demonstrator of \cref{app:clean_demo}. We state this condition in \cref{sec:res_drills}.

\begin{table}[htbp]
\centering\footnotesize\setlength{\tabcolsep}{3pt}

\caption{\textbf{How much of the full gain rehearsal alone reaches: target-harness score (\%)}}
\label{tab:reb_G}
\begin{tabular}{@{}lcccccc@{}}
\toprule
 & \makecell{Rehearsal\\only} & \makecell{Task +\\rehearsal} & Specialist & 3H SFT & \makecell{Rehearsal /\\specialist} & \makecell{Rehearsal /\\3H SFT} \\
\midrule
\makecell[l]{BFCL multi-turn,\\\textsc{fc}} & 32.5 & 58.8 & 55.2 & 59.0 & 59\% & 55\% \\
\addlinespace[3pt]
\makecell[l]{Reasoning Gym easy,\\Terminus-2} & 20.0 $\pm$ 5.7 & 37.0 $\pm$ 7.1 & 54.7 $\pm$ 7.8 & 55.3 $\pm$ 7.9 & 37\% & 36\% \\
\addlinespace[3pt]
\makecell[l]{MedAgentBench,\\Terminus-2} & 9.5 $\pm$ 5.7 & 34.4 $\pm$ 8.2 & 21.7 $\pm$ 6.4 & 16.9 $\pm$ 5.9 & 44\% & 56\% \\
\addlinespace[3pt]
\makecell[l]{MedAgentBench,\\\textsc{fc}} & 39.6 $\pm$ 9.6 & 36.6 $\pm$ 9.8 & 45.1 $\pm$ 9.4 & 55.3 $\pm$ 8.2 & 88\% & 72\% \\
\bottomrule
\end{tabular}\par

\end{table}

\subsection{Task Leakage Through Rehearsal}
\label[appendix]{app:leakage}
MAB has 10 task templates, all represented in the test set, and rehearsal was built from 4 of them (task2, task3, task4 and task8), so six test templates were never rehearsed. No training task appears in the test set. If the gain came from leaked task knowledge it should concentrate on the rehearsed templates. \cref{tab:reb_I} pools the templates and \cref{tab:reb_M} reports each one. All six never-rehearsed templates improve (+13.9 to +44.4). On them, task supervision alone (8.5) and rehearsal alone (9.1) remain weak while their combination reaches 42.2, so the complementarity is fully present on templates the model never rehearsed. The two rehearsed read templates gain more (+80.5 and +58.2), as expected when rehearsal practises the same request patterns the test tasks use and since they are among the easiest templates overall. The two rehearsed write templates gain little (+1.4 and +6.9) and are the hardest templates on the Terminus-2 target.

\begin{table}[htbp]
\centering\footnotesize\setlength{\tabcolsep}{4pt}

\caption{\textbf{Leakage analysis on MedAgentBench: success (\%) on rehearsed vs never-rehearsed templates}}
\label{tab:reb_I}
\begin{tabular}{@{}lcc@{}}
\toprule
 & \makecell{Rehearsed templates\\(task2/3/4/8)} & \makecell{Never-rehearsed templates\\(task1/5/6/7/9/10)} \\
\midrule
Task supervision only & 5.9 $\pm$ 4.0 & 8.6 $\pm$ 4.8 \\
Task supervision + rehearsal & \textbf{42.5 $\pm$ 15.6} & \textbf{42.6 $\pm$ 8.7} \\
Gain of the combined arm (derived) & +36.6 & +34.0 \\
\bottomrule
\end{tabular}\par

\end{table}
\begin{table}[htbp]
\centering\footnotesize\setlength{\tabcolsep}{4pt}

\caption{\textbf{MedAgentBench success (\%) per test template, Terminus-2, mean of three training seeds} Gain: task + rehearsal minus task only. 6 test tasks per template (task6: 7), 4 attempts each. The mean row weights templates equally, \cref{tab:reb_I} weights tasks.}
\label{tab:reb_M}
\begin{tabular}{@{}lccccc@{}}
\toprule
Template & Rehearsed & Task only & Rehearsal only & Task + rehearsal & Gain \\
\midrule
task2 & yes (read) & 5.6 & 65.1 & 86.1 & +80.5 \\
task4 & yes (read) & 18.2 & 8.9 & 76.4 & +58.2 \\
task3 & yes (write) & 0.0 & 4.3 & 1.4 & +1.4 \\
task8 & yes (write) & 0.0 & 1.4 & 6.9 & +6.9 \\
task1 & no (read) & 1.4 & 9.7 & 40.3 & +38.9 \\
task5 & no (read + conditional write) & 33.3 & 26.1 & 73.6 & +40.3 \\
task6 & no (read) & 10.8 & 0.0 & 53.0 & +42.2 \\
task7 & no (read) & 0.0 & 4.2 & 22.6 & +22.6 \\
task9 & no (read + conditional write) & 4.2 & 13.1 & 48.6 & +44.4 \\
task10 & no (read + conditional write) & 1.4 & 1.4 & 15.3 & +13.9 \\
Never rehearsed, mean &  & 8.5 & 9.1 & 42.2 & +33.7 \\
\bottomrule
\end{tabular}\par

\end{table}

\subsection{Rehearsal from a Clean Demonstrator}
\label[appendix]{app:clean_demo}
\cref{tab:reb_J} repeats the $2\times2$ with rehearsal traces from a demonstrator that never saw target-harness solutions (the Qwen3-32B base model). The combined arm keeps its gain: within about three points of the original-demonstrator arm on BFCL-MT and MAB, and equal or higher on RG, where rehearsal alone from this demonstrator is already strong (48.8 / 36.6) and task supervision adds little.

\begin{table}[htbp]
\centering\footnotesize\setlength{\tabcolsep}{4pt}

\caption{\textbf{Clean demonstrator (Qwen3-32B base, never trained with target-harness solutions): target-harness score (\%)}}
\label{tab:reb_J}
\begin{tabular}{@{}lcccc@{}}
\toprule
 & \makecell{Rehearsal only,\\3H demonstrator} & \makecell{Rehearsal only,\\clean (32B)\\demonstrator} & \makecell{Task + rehearsal,\\3H demonstrator} & \makecell{Task + rehearsal,\\clean demonstrator} \\
\midrule
\makecell[l]{BFCL multi-turn\\(\textsc{fc})} & 33.6 $\pm$ 5.1 & 30.5 $\pm$ 5.2 & 57.5 $\pm$ 5.6 & 54.2 $\pm$ 5.6 \\
\addlinespace[3pt]
\makecell[l]{MedAgentBench\\(Terminus-2)} & 12.9 $\pm$ 7.2 & 14.4 $\pm$ 6.6 & 35.2 $\pm$ 8.4 & 32.0 $\pm$ 9.0 \\
\addlinespace[3pt]
\makecell[l]{Reasoning Gym easy\\(Terminus-2)} & 21.3 $\pm$ 5.7 & 48.8 $\pm$ 7.8 & 43.7 $\pm$ 7.5 & 44.4 $\pm$ 7.6 \\
\addlinespace[3pt]
\makecell[l]{Reasoning Gym hard\\(Terminus-2)} & 17.7 $\pm$ 5.1 & 36.6 $\pm$ 8.1 & 30.4 $\pm$ 6.5 & 32.6 $\pm$ 7.1 \\
\bottomrule
\end{tabular}\par

\end{table}

\subsection{What Rehearsal Changes: Failure Types}
\label[appendix]{app:failure_types}
To see how rehearsal changes failure modes, we analysed failed trials from the task-supervision-only and task + rehearsal checkpoints on each benchmark's target harness, using one checkpoint per condition and 4 attempts per task: 3,618 trials in total, of which 2,347 failed. Automatic labels cover all failed trials. For a human analysis we sampled 100 failures with a fixed seed, 50 per condition, split evenly across the three benchmarks. Three annotators, blind to checkpoint identity but shown the automatic label, classified each failure as a task-knowledge failure (well-formed actions, incorrect content) or one of three interface failure types: \emph{invalid calls} (the harness could not parse or execute the action), \emph{incorrect submissions} (the answer was not delivered as the harness expects) and \emph{termination failures} (the run did not end properly). Each trial was labelled by its first decisive error. The automatic rules are deterministic but imperfect: a trial whose answer reaches the verifier and is judged wrong is a task failure. Otherwise it is an invalid call if at least half of its turns were re-prompted or any call used another harness's syntax, a termination failure if it timed out or ended on a re-prompt, and an incorrect submission otherwise. The main disagreement with the annotators occurs on MAB after rehearsal, where some runs write an answer and then loop on invalid calls until timeout, which the rules classify as task failures and annotators as interface failures (\cref{tab:reb_F}). Annotator agreement is Fleiss $\kappa=0.61$, and all three agree on 80\% of trials. Rehearsal reduces failure rates on every benchmark. On BFCL-MT and RG, interface failures fall by more than the total gain in success, so runs that previously failed at the harness now proceed far enough to fail on task content instead. On MAB, failures drop from 223/244 to 158/244 and 19.1 of the 26.6 percentage-point gain comes from fewer interface failures, while the remaining interface errors shift from incorrect submissions (82\%$\to$41\% of failures) to invalid calls (4\%$\to$41\%) (\cref{tab:reb_L}). The failure analysis therefore supports the reading that rehearsal primarily improves reliable use of the target interface.

\begin{table}[htbp]
\centering\footnotesize\setlength{\tabcolsep}{4pt}

\caption{\textbf{Interface failures, automatic vs human} Annotator agreement (interface vs task knowledge): Fleiss $\kappa$ = 0.61, and all three agree on 80\% of trials.}
\label{tab:reb_F}
\begin{tabular}{@{}llccc@{}}
\toprule
 &  & \makecell{Failed trials\\(\% of all)} & \makecell{Interface failures,\\\% of all trials\\(auto $\cdot$ human)} & \makecell{Interface failures,\\\% of failed trials\\(auto $\cdot$ human)} \\
\midrule
\makecell[l]{BFCL multi-turn\\(\textsc{fc})} & Task supervision only & 76.2 & 65.8 $\cdot$ 71.5 & 86 $\cdot$ 94 \\
 & \textbf{Task + rehearsal} & \textbf{42.1} & \textbf{7.0 $\cdot$ 10.5} & \textbf{17 $\cdot$ 25} \\
\makecell[l]{Reasoning Gym\\(Terminus-2)} & Task supervision only & 78.1 & 69.7 $\cdot$ 70.5 & 89 $\cdot$ 90 \\
 & \textbf{Task + rehearsal} & \textbf{55.1} & \textbf{38.2 $\cdot$ 38.9} & \textbf{69 $\cdot$ 71} \\
\makecell[l]{MedAgentBench\\(Terminus-2)} & Task supervision only & 91.4 & 73.8 $\cdot$ 78.8 & 81 $\cdot$ 86 \\
 & \textbf{Task + rehearsal} & \textbf{64.8} & \textbf{25.4 $\cdot$ 59.7} & \textbf{39 $\cdot$ 92} \\
\bottomrule
\end{tabular}\par

\end{table}
\begin{table}[htbp]
\centering\footnotesize\setlength{\tabcolsep}{4pt}

\caption{\textbf{Interface failure modes, \% of failed trials (automatic $\cdot$ human)} Automatic percentages use all failures, and human percentages use sampled failures. The remainder of each row is task-knowledge failures.}
\label{tab:reb_L}
\begin{tabular}{@{}llccc@{}}
\toprule
 &  & \makecell{Invalid\\calls} & \makecell{Incorrect\\submissions} & \makecell{Termination\\failures} \\
\midrule
\makecell[l]{BFCL multi-turn\\(\textsc{fc})} & Task supervision only & 85 $\cdot$ 88 & n/a $\cdot$ 4 & 2 $\cdot$ 2 \\
 & \textbf{Task + rehearsal} & \textbf{4 $\cdot$ 15} & \textbf{n/a $\cdot$ 0} & \textbf{12 $\cdot$ 10} \\
\makecell[l]{Reasoning Gym\\(Terminus-2)} & Task supervision only & 74 $\cdot$ 69 & 6 $\cdot$ 14 & 9 $\cdot$ 8 \\
 & \textbf{Task + rehearsal} & \textbf{5 $\cdot$ 8} & \textbf{52 $\cdot$ 60} & \textbf{12 $\cdot$ 2} \\
\makecell[l]{MedAgentBench\\(Terminus-2)} & Task supervision only & 3 $\cdot$ 4 & 76 $\cdot$ 82 & 2 $\cdot$ 0 \\
 & \textbf{Task + rehearsal} & \textbf{38 $\cdot$ 41} & \textbf{1 $\cdot$ 41} & \textbf{1 $\cdot$ 8} \\
\bottomrule
\end{tabular}\par

\end{table}

\section{Limitations and Future Work}
\label[appendix]{sec:limitations}

We evaluate one held-out harness per benchmark rather than exhaustively covering all possible interfaces. A useful extension would be to study a broader range of held-out harnesses and relate transfer to their similarity to the training harnesses. In particular, an explicit measure of harness distance could help predict when cross-harness supervision will transfer and how much target-interface practice is required.

Our primary goal is to study how task and interface competence interact during training, rather than to benchmark the effectiveness of a particular training method across model families. We therefore focus on Qwen3-4B, with additional results on Gemma-3-4B-it (\cref{app:gemma}). The main training-harness patterns are similar across both models, while differences in held-out transfer suggest that sensitivity to interface training may vary across model families.

We focus on tasks that can be executed across multiple harnesses without relying on specific tool availability, external memory, or skills. This helps isolate the general effect of different harnesses, but excludes settings in which harnesses differ in the tools or auxiliary capabilities they expose. Extending the analysis to such settings would help disentangle interface competence from competence that depends on access to additional resources.

\end{document}